\documentclass[letterpaper]{article} % DO NOT CHANGE THIS
\usepackage{aaai2026}  % DO NOT CHANGE THIS
\usepackage{times}  % DO NOT CHANGE THIS
\usepackage{helvet}  % DO NOT CHANGE THIS
\usepackage{courier}  % DO NOT CHANGE THIS
\usepackage[hyphens]{url}  % DO NOT CHANGE THIS
\usepackage{graphicx} % DO NOT CHANGE THIS
\usepackage{natbib}  % DO NOT CHANGE THIS AND DO NOT ADD ANY OPTIONS TO IT
\usepackage{caption} % DO NOT CHANGE THIS AND DO NOT ADD ANY OPTIONS TO IT
\usepackage{algorithm}
\usepackage{amsmath}
\usepackage{algpseudocode}
\usepackage{booktabs}
\usepackage{multirow}

\nocopyright

\usepackage{newfloat}
\usepackage{listings}
\DeclareCaptionStyle{ruled}{labelfont=normalfont,labelsep=colon,strut=off} % DO NOT CHANGE THIS
\floatstyle{ruled}
\newfloat{listing}{tb}{lst}{}
\floatname{listing}{Listing}
\title{Simple o3: Towards Interleaved Vision-Language Reasoning}
\author {
    Ye Wang\textsuperscript{\rm 1},
    Qianglong Chen\footnotemark[1],
    Zejun Li\textsuperscript{\rm 1}, 
    Siyuan Wang\textsuperscript{\rm 2}, \\
    Shijie Guo\textsuperscript{\rm 1},
    Zhirui Zhang\footnotemark[1],
    Zhongyu Wei\textsuperscript{\rm 1}
}
\affiliations {
    \textsuperscript{\rm 1}Fudan University\\
    \textsuperscript{\rm 2}University of Southern California\\
    yewang22@m.fudan.edu.cn,\\
    {\{chenqianglong.ai, zrustc11\}@gmail.com}, \\
    {\{zejunli20, guoshijie, zywei\}@fudan.edu.cn}, \\
    {sw\_641@usc.edu}
}

\usepackage{bibentry}
\begin{document}

\maketitle
\renewcommand{\thefootnote}{*}
\footnotetext[1]{Independent Researcher}

\begin{abstract}
Multimodal Large Language Models (MLLMs) have shown impressive performance on vision-language tasks, but their long Chain-of-Thought (CoT) capabilities in multimodal scenarios remain underexplored. Inspired by OpenAI's o3 model, which emulates human-like ``thinking with image'' through iterative visual transformations and linguistic reasoning, we propose Simple o3, an end-to-end framework that integrates dynamic tool interactions (e.g., cropping, zooming, and reusing) into interleaved vision-language reasoning via supervised fine-tuning (SFT). Our approach features a scalable data synthesis pipeline that generates high-quality interleaved vision-language reasoning chains via an ``observe-reason-act'' cycle, complete with executable visual operations and rigorous verification, yielding the open-source TWI-Tools-146K dataset. Experimental results demonstrate Simple o3's  superior performance on diverse benchmarks, outperforming existing approaches. By combining enhanced reasoning capabilities, Simple o3 establishes a powerful yet computationally affordable paradigm for advancing multimodal reasoning. Remarkably, we provide the first in-depth analysis of different interleaved reasoning strategies, offering insights into their impact on model performance. We found that by introducing additional visual tokens for interleaved vision-language reasoning, reusing and magnifying the original image significantly improves the model's visual reasoning and fine-grained perception, while image cropping based on precise visual grounding allows the model to effectively focus on key entities or regions, further enhancing its capabilities.
\end{abstract}

\section{Introduction}
Recent advancements in Multimodal Large Language Models (MLLMs) have demonstrated remarkable proficiency across a wide range of vision-language tasks~\cite{bai2025qwen2,chen2024expanding,liu2024improved,li2024llava,team2025kimi}, from image captioning to visual question answering (VQA). Concurrently, the exploration into ``slow-thinking'' capabilities in large language models (LLMs), inspired by human deliberative reasoning processes, has gained traction.~\cite{wei2022chain,kojima2022large,guo2025deepseek,yu2025dapo}. 

However, research on extended Chain-of-Thought (CoT) in multimodal scenarios—particularly those involving interleaved visual-language reasoning pathways—remains largely underexplored. A seminal advance in this direction is OpenAI’s newly proposed O3 model, which emulates human-like reasoning by ``thinking with images''—an iterative paradigm where visual perception and cognitive processing co-evolve. At each reasoning step, the model transforms images (e.g., by cropping, zooming, or rotating), concurrently performing linguistic reasoning to plan subsequent visual operations. This forms a groundbreaking framework for multimodal agentic reasoning, bridging continuous visual and discrete linguistic reasoning. While tool-augmented reasoning has demonstrated promise in language-only domains~\cite{li2025torl,jin2025search}, there lacks a systematic methodology to activate such capabilities in multimodal scenarios that could enable hierarchical task decomposition, akin to how humans revisit and manipulate visual stimuli during complex problem-solving. Second, current approaches for eliciting tool-use behaviors in MLLMs rely heavily on resource-intensive Reinforcement Learning (RL) or human annotation, lacking scalable pipelines for synthesizing high-quality interleaved vision-language reasoning data~\cite{liu2025visionreasoner,fan2025gritteachingmllmsthink,liu2025seg}. Moreover, which tools can effectively enhance multimodal reasoning capabilities and how input resolution affects interleaved vision-language reasoning remain underexplored. Addressing this gap is key to optimizing visual information usage and advancing model reasoning capabilities.

\begin{figure*}[htbp]
    \centering
    \includegraphics[width=.95\textwidth]{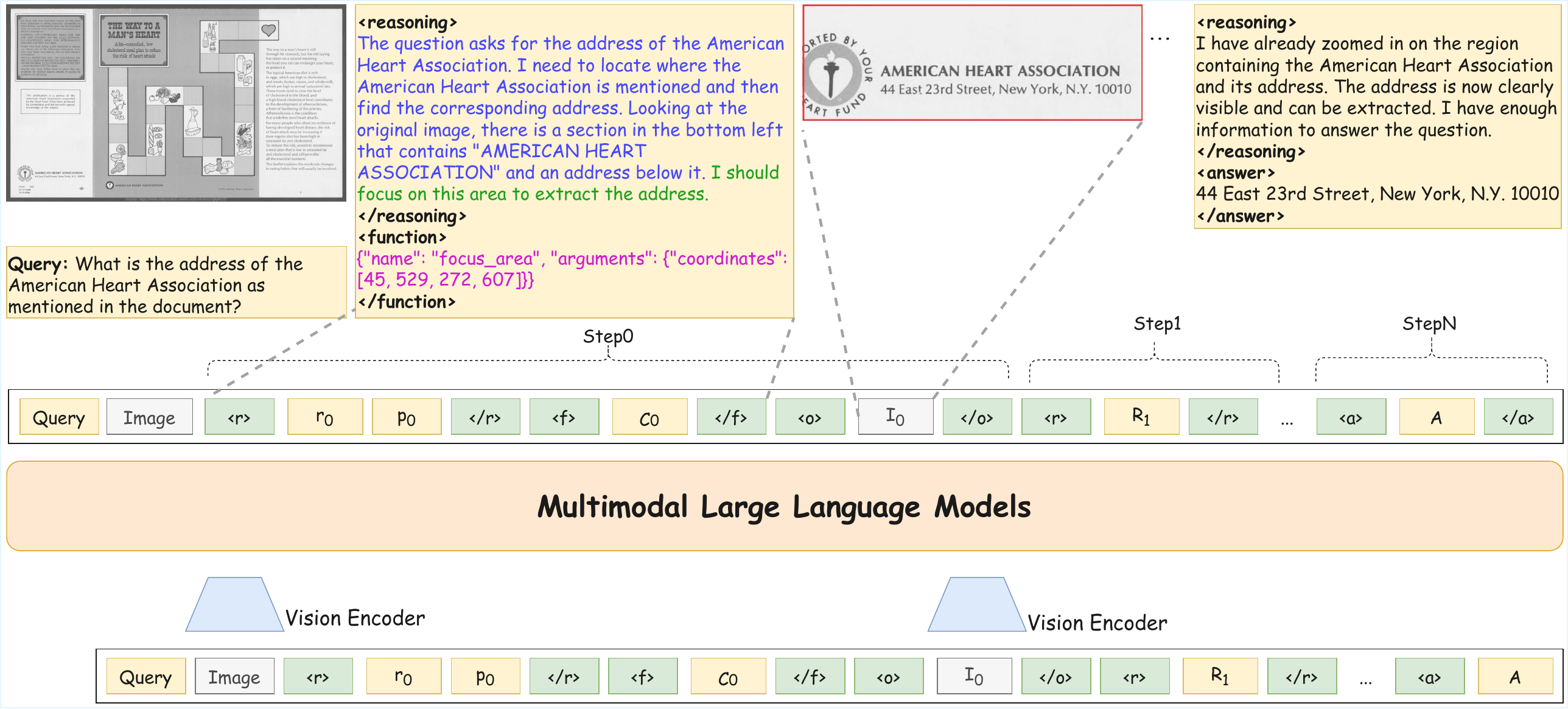}
    \caption{Overview of Simple o3. At Step 0, the blue text represents the atomic reasoning step $r_0$, while the green text represents the visual operation plan $p_0$. These two components constitute the reasoning content $R_0$. The pink text  represents the tool instruction $C_0$, which is returned as JSON object. During training, $focus\_area$ operation is followed by an image with the bbox to obtain the complete visual information of the image. During inference, the image is cropped according to the coordinates returned by $focus\_area$ to inject visual tokens of target entities or regions.}
    \label{fig:simple_o3}
\end{figure*}

In this paper, we introduce Simple o3, a simple yet powerful open-source framework for end-to-end learning of tool-augmented reasoning with interleaved image-text inputs. The framework combines a scalable data generation pipeline, a training method that incorporates image masking, and a multi-step reasoning enabling seamless tool interaction, offering an integrated solution for complex multimodal tasks. Specifically, we propose a scalable, high-quality data synthesis pipeline that automatically generates interleaved vision-language reasoning chains through an iterative ``observe-reason-act'' cycle. Given a question-image pair, the pipeline first produces atomic reasoning steps alongside corresponding visual operation plannings, then converts these plannings into executable parameters for precise image transformations (e.g., $focus\_area$, $zoom\_in$, or $reuse$). Using designated visual tools, the system applies these parameters to update the visual state, iteratively repeating the cycle until the accumulated reasoning steps provide sufficient evidence to trigger the answer. Meanwhile, the two-step verification module ensures semantic consistency between visual operation planning and output parameters while filtering out answers inconsistent with the ground truth, thereby maximizing the retention of high-quality data. During training, we reformat the multi-turn dialogue data into a single-turn user-assistant structure, unify coordinate systems, and apply image masking for loss computation. During inference, the model detects parameters within function tags at each step, executes the corresponding image operations, and iterates until the answer is derived. This tool-interactive approach to interleaved image-text reasoning substantially enhances the model's logical capabilities, as demonstrated in Figure~\ref{fig:simple_o3}.

Our main contributions are fourfold:
(1) We propose Simple o3, which reproduces and extends the o3 model's tool-interactive ``thinking with images'' paradigm, advancing the development of multimodal CoT reasoning.
(2) We introduce an iterative, high-quality data synthesis pipeline, providing detailed methodology for generating image-text interleaved reasoning data with executable function calls, accompanied by the release of our open-source TWI-Tools-146K dataset.
(3) Experimental results demonstrate that Simple o3 achieves significant improvements on multimodal reasoning and perception benchmarks, validating the effectiveness of our approach.
(4) We present the first in-depth analysis of the impact of tool selection and input resolution on interleaved vision-language reasoning performance, providing key insights for future ``thinking with images'' research.

\section{Related Work}

\subsection{Multi Large Language Models}
By integrating visual encoders with powerful LLMs, MLLMs gain visual perception capabilities~\cite{alayrac2022flamingo,awadalla2023openflamingo}. Early MLLMs primarily followed architectures like LLaVA~\cite{liu2024visual} and BLIP-2~\cite{li2023blip}, employing either MLP-based or attention-based connector modules to bridge the visual and language modalities~\cite{zhu2023minigpt,bai2023qwen,chen2023shikra}. In these frameworks, the visual encoder processes an input image, after which the connector module projects the visual features to align dimensionally with text embeddings, forming a unified multimodal sequence representation. Since most visual encoders are constrained by fixed input resolutions and capturing only partial visual features, subsequent research has prioritized enhancing visual representations. Two predominant approaches have emerged: some directly extend the visual encoders' capabilities~\cite{yuan2024osprey,ge2024convllava,fan2024mousi}, and others adopt dynamic resolution strategies that process high-resolution images by decomposing them into multiple sub-images while simultaneously analyzing downsampled global views through low-resolution encoders~\cite{ye2023ureader,li2024monkey,li2024llava,chen2024expanding}. While these innovations have substantially improved visual perception, the persistent limitations in reasoning capacity continue to constrain overall model performance.

\subsection{Long CoT in MLLMs}
Early MLLMs predominantly adopted a two-stage training paradigm consisting of vision-language pretraining followed by instruction fine-tuning~\cite{dai2023instructblip,liu2023visual,ye2023mplug,zeng2024matters}. While this approach improved instruction-following capabilities, it inherently separates perception from reasoning. Subsequent research has sought to address this limitation by integrating various forms of textual auxiliary inputs, including detailed reasoning traces, spatial coordinates, attribute relations, and comprehensive image descriptions, to provide evidential grounding for model outputs~\cite{shao2024visual,xu2024llava}. The emergence of DeepSeek-R1 marked a turning point where RL became a key enabler for reasoning enhancement, spawning diverse training paradigms. Similar advancements soon appeared in multimodal domains. While recent research has increasingly focused on reward signal design~\cite{huang2025vision,liu2025visual,zhang2025r1,zhou2025r1}, OpenAI's o3 model pioneered the 'thinking with images' paradigm and has gradually become pivotal for enhancing multimodal reasoning capabilities~\cite{zhang2025cmmcot,wang2025vrag,bai2025multi,ni2025point,chung2025don}. Approaches like DeepEyes~\cite{zheng2025deepeyes} and Chain-of-Focus~\cite{zhang2025chain} dynamically leverage image cropping tools to facilitate adaptive visual reasoning~\cite{zhu2025active,su2025pixel}. However, these works constrain the reasoning process to basic image cropping operations. This limited toolset results in rigid inference pipelines that lack adaptability to diverse scenarios, where different tasks may require distinct visual processing strategies. More critically, existing work fails to systematically examine how tool selection and input resolution influences fundamental model capabilities, particularly in perceptual understanding and reasoning.

\section{Simple o3}

In this section, we introduce Simple o3 through three key components: (1) the scalable data generation pipeline that automates high-quality reasoning chain creation, (2) the novel training methodology incorporating image masking, and (3) the multi-step inference enabling dynamic tool interaction. These components collectively form an end-to-end system for interleaved image-text reasoning, achieving robust performance through adaptive multi-tool collaboration.

\subsection{Scalable Data Generation Pipeline}
We propose a data generation pipeline consisting of a reasoning path generator and a two-step verification module, as shown in Figure~\ref{fig:generation_pipeline}. In the generation phase, the MLLM serves as an orchestrator that dynamically analyzes task requirements and coordinates predefined visual tools: $focus\_area$ returns the image with a drawn bounding box (bbox) of a focused region, $zoom\_in$ magnifies the whole image area via interpolation, and $reuse$ directly outputs the original image. Each tool encapsulates a dedicated image processing  function, which can perform the atomic manipulation on the input image. We employ \emph{gemini-2.5-Flash-Preview-05-20} in its non-thinking mode to optimize the trade-off between generation efficiency and computational overhead. The reasoning path generation follows an iterative cycle:

\begin{equation}
S = \{s_0, s_1, ..., s_t, A\}
\end{equation}

\begin{equation}
s_t = (R_t, C_t, I_t) \sim P(\cdot \mid Q, I, H_{t-1};\theta_{\text{MLLM}}) \quad (t \geq 0)
\end{equation}

% \begin{equation}
% H_{t-1} = \{(R_{i}, C_{i}, I_{i})\}^{i=t-1}_{i=0} \quad (t \geq 1)
% \end{equation}
\begin{equation}
H_{t-1} = 
   \begin{cases} 
   \emptyset & \text{if } t = 0, \\
   \{(R_i, C_i, I_i)\}_{i=0}^{t-1} & \text{if } t \geq 1.
   \end{cases}
\end{equation}

Here, $S$ represents the iterative reasoning chain comprising the serialized atomic reasoning step $s$, and final answer $A$. $s_t$ denotes the output at reasoning step $t$, including the reasoning content $R_t$, the tool command $C_t$, and the returned image $I_t$, formed as the triple, by sampling from the MLLM while considering the historical reasoning path $H_{t-1}$ that aggregates all previous steps' information as the reasoning process unfolds.

% Our pipeline utilizes structured tags as special tokens to delineate distinct reasoning components: atomic reasoning steps $r_t$ and visual planning $p_t$ constitute the text output $R_t = (r_t, p_t)$, are enclosed within \emph{$<$reasoning$>$} and \emph{$<$/reasoning$>$} tags, while tool invocation details $C_t = (T_n, Param)$, including both the operation name and required input parameters, are contained within \emph{$<$function$>$} and \emph{$<$/function$>$} tags. The system dynamically generates new visual states through these operations, with the resulting transformed images being encoded and inserted between \emph{$<$observation$>$} and \emph{$<$/observation$>$} delimiters, thereby maintaining a complete and parseable record of the entire reasoning chain's visual-textual evolution.

Our pipeline utilizes structured tags as special tokens to delineate distinct reasoning components. The reasoning content $R_t = (r_t, p_t)$, comprising atomic reasoning steps $r_t$ and visual planning $p_t$, is encapsulated within \emph{$<$reasoning$>$} and \emph{$<$/reasoning$>$} tags. Tool invocations $C_t = (T_n, Param)$, specifying both operation names and input parameters, are marked by \emph{$<$function$>$} and \emph{$<$/function$>$} tags. The system dynamically generates new visual states through image manipulations, with the resulting transformed images being encoded and embedded between \emph{$<$observation$>$} and \emph{$<$/observation$>$} delimiters. This structured tagging scheme preserves a complete and parseable record of multimodal reasoning chain progression.

% In the verification module, the first step is to check the geometric rationality of all tool operation executions, using \emph{gemini-2.5-flash-lite-preview-06-17}. For the image key region focusing operation, samples where the bounding box (bbox) matches or contains the target region/entity are retained. For zoom-in and reuse operations, samples where $C_t$ is semantically consistent with the visual operation planning $p_t$ are kept. Finally, \emph{Qwen3-turbo} is used to compare the output answer with the ground truth. For each data sample, we provide the image, question, and answer to the MLLM to generate a multi-step reasoning chain based on tool invocation. For each question-image pair, the system makes a maximum of two generation attempts to increase valid samples.

The verification process begins by assessing the geometric validity of tool interactions using \emph{gemini-2.5-flash-lite-preview-06-17}. For $focus\_area$, we retain samples where the bbox either precisely matches or fully contains the target entity. $zoom\_in$ and $reuse$ operations are validated based on semantic consistency between the tool command $C_t$ and the visual planning description $p_t$. Finally, the answer is verified by comparing system outputs against ground truth using \emph{Qwen3-turbo}. For each question-image-answer triplet, the MLLM constructs a multi-step reasoning chain, potentially invoking visual tools. The system performs a maximum of two generation attempts per sample to enhance validity.

\subsection{TWI-Tools-146K Data Curation}
% We carefully curate the dataset from diverse tasks to maximize the data distribution, thereby enhancing Simple o3's generalization capability. Our TWI-Tools-146K training dataset is primarily sourced from the following datasets: MATHV360K~\cite{shi2024math}, LLaVA-CoT-100K~\cite{xu2024llava}, and ArxivQA~\cite{li2024multimodal}. We excluded subtasks including math problems from MATHV360K, as geometry often requires consideration of the entire image, and proportionally sample 150K entries proportionally across its remaining sub-datasets to generate ``thinking with images'' data. After filtering through a verification module, approximately 100K high-quality entries were retained, covering perceptual visual QA, knowledge-based reasoning, chart data, and logical reasoning, representing a collection of challenging everyday questions. To establish a solid foundation for visual understanding and generalization, we mainly select COCO QA pairs from LLaVA-CoT-100K and generate 27K samples, which feature rich real-world images. To further strengthen the model’s ability to interpret and reason with abstract charts, we extracted 19K entries from ArxivQA proportionally by subcategory.

We carefully curate the dataset from diverse tasks to maximize the data distribution, thereby enhancing Simple o3's generalization capability. Our TWI-Tools-146K training dataset is sourced from following datasets: MATHV360K~\cite{shi2024math}, LLaVA-CoT-100K~\cite{xu2024llava}, and ArxivQA~\cite{li2024multimodal}. We exclude subtasks including math problems from MATHV360K, as geometry typically focuses more on the extension of textual tokens. Through proportional sampling across remaining sub-datasets, we initially generate 150K candidate entries. After rigorous verification, we retain approximately 100K high-quality samples, covering perceptual VQA, knowledge-based reasoning, chart analysis, and logical reasoning, representing a comprehensive collection of challenging real-world tasks. To establish robust visual understanding capabilities, we primarily select COCO QA pairs from LLaVA-CoT-100K, generating 27K samples featuring diverse real-world imagery. To further enhance abstract figure interpretation and reasoning,, we extract 19K proportionally balanced samples from ArxivQA, covering all subcategories.

% Overview of the scalable vision-language interleaved reasoning data generation pipeline: MLLM serves as the decision - maker, first understanding the task and decomposing it into multi - step reasoning. In each step, MLLM generates the reasoning content of the current step and tool instructions. The toolbox, which contains three functions namely $focus\_area$, $zoom\_in$ and $reuse$, processes the input image according to the invocation instructions and returns the image after visual operations. The tool verification, specifically judging whether the tool - invocation result is semantically aligned with the visual operation plan, conducts the first - step check. After passing the check, the reasoning content, tool - invocation instructions, and the post - operation image are concatenated as historical information and input into MLLM for the next - step generation until an answer is output. After going through the answer verification module, a complete sample is obtained.

\begin{figure}[htbp]
    \centering
    \includegraphics[width=0.45\textwidth]{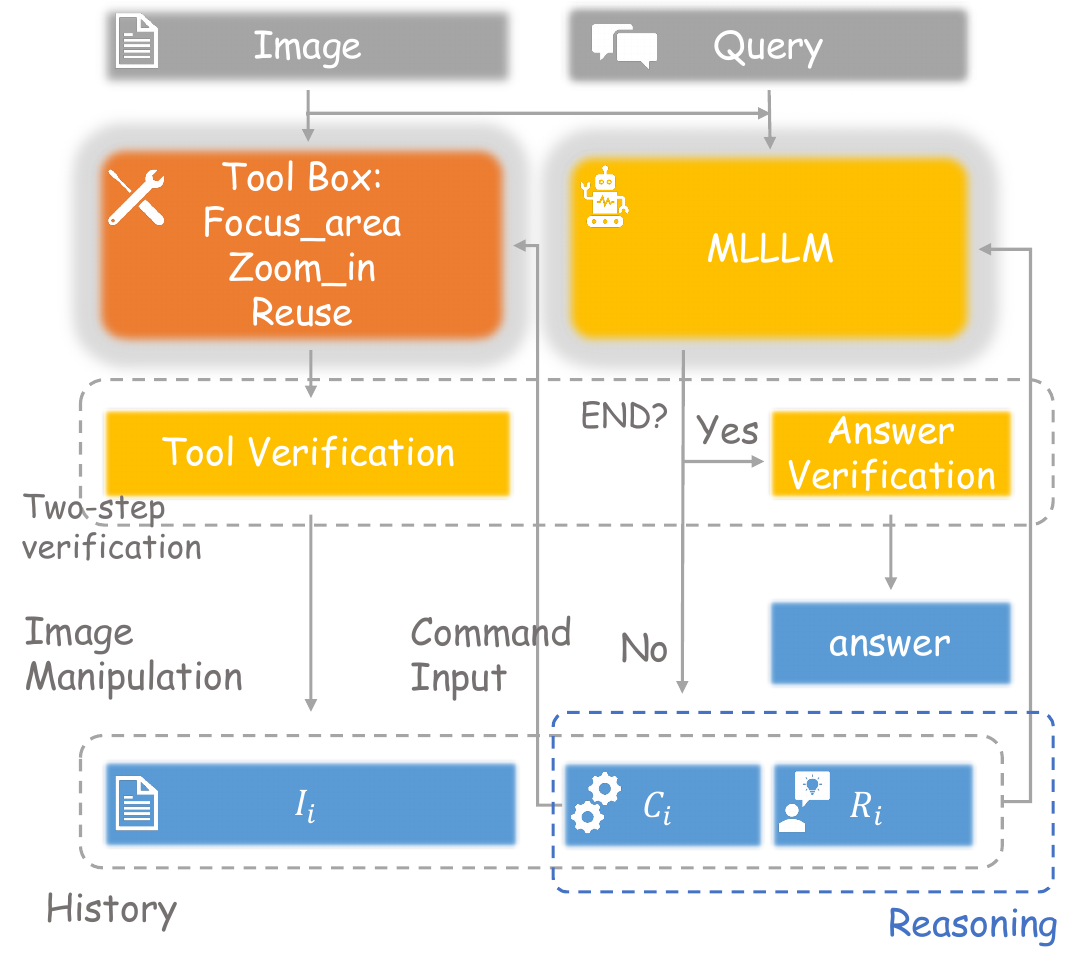}
    \caption{Overview of scalable data generation pipeline. MLLM generates current reasoning content and tool commands at each step. The toolbox processes input images based on these commands, returning manipulated visuals, followed by tool verification to ensure semantic alignment between commands and visual operations. Upon successful verification, the system combines the current step's generation into the history, for the next generation step. This cycle repeats until the answer is generated, concluding with answer verification to produce a complete data sample.}
    \label{fig:generation_pipeline}
\end{figure}

% \begin{figure}[htbp]
%     \centering
%     \includegraphics[width=0.45\textwidth]{AuthorKit26/AnonymousSubmission/LaTeX/prompt.pdf}
%     \caption{Overview of the scalable data generation pipeline.}
%     \label{fig:prompt}
% \end{figure}

% \subsection{Training Incorporating Image Masking}
% Through the data generation pipeline, we obtain series of interleaved vision-language sequences $\mathbf{Z} = (Q, I, S)$, where the multi-modal sequence can be further decomposed into the individual element $z_t$ from different modalities, originating from the text space $\mathcal{T}$ or the visual space $\mathcal{V}$. However, since the generated data and trained model may use incompatible coordinate systems, the coordinates must first be converted to the model's compatible format. For a multimodal sequence $\mathbf{Z} = (z_0, \dots, z_t)$, we introduce a selective loss computation using a modality mask $M_t$:

\subsection{Training Incorporating Image Masking}
% Through the data generation pipeline, we obtain series of interleaved vision-language sequences $\mathbf{Z} = (Q, I, S)$, where the iterative reasoning chain $S$ can be further decomposed into the individual element $z_t$ from different modalities, originating from the text space $\mathcal{T}$ or the visual space $\mathcal{V}$. However, since the generated data and trained model may use incompatible coordinate systems, the coordinates must first be converted to the model's compatible format. For a multimodal reasoning sequence $S = (z_0, \dots, z_t)$, we introduce a selective loss computation using a modality mask $M_t$:
Our data generation pipeline produces interleaved vision-language sequences $\mathbf{Z} = (Q, I, S)$, where the iterative reasoning chain $S$ can be further decomposed into individual elements $z_t$ from different modalities, originating from the text space $\mathcal{T}$ or the visual space $\mathcal{V}$. To adapt the multi-turn ``user-assistant-tool'' dialogues obtained during data generation into a suitable training format, we consolidate their reasoning paths into single-turn ``user-assistant'' sequences. Prior to training, we also resolve potential coordinate system discrepancies by converting all spatial coordinates to the native representation format of the target model. For effective learning from multimodal reasoning sequences $S = (z_0, \dots, z_t)$, we employ a selective loss computation mechanism governed by modality-aware masking $M_t$, which ensures proper gradient flow while maintaining cross-modal interactions.

% Since we obtain multi-turn 'user-assistant-tool' dialogues during the data generation phase, we need to concatenate their reasoning paths into single-turn user-assistant sequences. To address potential coordinate system incompatibilities between generated data and the model's requirements, we first transform all coordinates into the model's native format. For training with multimodal reasoning sequences $S = (z_0, \dots, z_t)$, we implement selective loss computation through a modality-aware masking $M_t$:

% \begin{equation}
% M_t = \begin{cases} 
% 1 & \text{if } z_t \in \mathcal{T} \quad \text{(compute loss)} \\
% 0 & \text{if } z_t \in \mathcal{V} \quad \text{(ignore loss)}
% \end{cases}
% \end{equation}

\begin{equation}
M_t = \begin{cases} 
1 & \text{if } z_t \in \mathcal{T} \quad \text{(compute loss)} \\
0 & \text{if } z_t \in \mathcal{V} \quad \text{(ignore loss)}
\end{cases} \quad (t \geq 0)
\end{equation}

The training objective employs a masked cross-entropy loss that optimizes textual generation, zeroing out gradients for visual tokens:

% \begin{equation}
% \mathcal{L}_{\text{SFT}} = - \sum_{t=1}^{T} M_t \cdot \log P(\cdot \mid Q, I, H_{t-1};\theta_{MLLM})
% \end{equation}

% \begin{equation}
% \mathcal{L}_{\text{SFT}} = - \sum_{t=0}^{T} M_t \cdot \log P(z_t \mid Q, I, H_{t-1}; \theta_{\text{MLLM}}),
% \end{equation}
% where the history $H_{t-1}$ is defined as:
% \begin{equation}
% H_{t-1} = \begin{cases} 
% \emptyset & \text{if } t = 0 \quad \text{(no history)}, \\
% \{z_i\}_{i=0}^{t-1} & \text{if } t \geq 1.
% \end{cases}
% \end{equation}

\begin{equation}
\mathcal{L}_{\text{SFT}} = - \sum_{t=0}^{T} M_t \cdot \log P(z_t \mid Q, I, H_{t-1}; \theta_{\text{MLLM}}),
\end{equation}
where the history $H_{t-1}$ is defined as:
\begin{equation}
H_{t-1} = \begin{cases} 
\emptyset & \text{if } t = 0, \\
\{z_i\}_{i=0}^{t-1} & \text{if } t \geq 1.
\end{cases}
\end{equation}

In our implementation, visual tokens are demarcated by special tokens like \emph{$<|$image\_pad$|>$} and serve only as contextual inputs. The binary mask ensures the model attends to visual data for understanding while restricting gradient updates to textual outputs.

\subsection{Multi-step Inference with Tool Interaction}
After supervised fine-tuning, the model demonstrates proficiency in processing structured formats, enabling tool usage and multi-step interleaved vision-language reasoning. This capability facilitates dynamic alternation between text and image token generation until reaching an answer or hitting the round limit. The workflow is depicted in Algorithm~\ref{alg:rollout_search}: At each reasoning step $t$, the model outputs the reasoning content between \emph{$<$reasoning$>$} and \emph{$<$/reasoning$>$} tags. Upon detecting the \emph{$<$/function$>$} token, the system automatically pairs it with the preceding \emph{$<$function$>$} tag and extracts the JSON-formatted tool command $C_t$. Note that during inference, $focus\_area$ crops the image based on the returned coordinates instead of outputting the full image with a drawn bbox. This command is then passed to a predefined function for visual processing, and the resulting image $I_t$ is embedded as visual tokens between \emph{$<$observation$>$} and \emph{$<$/observation$>$} tags before being appended to the model’s history. Finally, the model performs summary reasoning and outputs the answer between \emph{$<$answer$>$} and \emph{$<$/answer$>$} tags, completing the inference cycle.

\begin{algorithm}[t]
\caption{MLLM Response with Multi-Turn Inference}
\label{alg:rollout_search}
\textbf{Require:} Input query \( Q \), Input image  \( I \), MLLM \( \pi_\theta \), tool box \( \mathcal{T} \), maximum rounds \( M \). \\
\textbf{Ensure:} Final response \( S \).
\begin{algorithmic}[1]
    \State Initialize multimodal reasoning sequence \( S \gets \emptyset \)
    \State Initialize round count \( r \gets 0 \)
    \While{$ r < M $}
        \State Generate response token \( y_t \sim \pi_\theta(\cdot \mid Q, I, S) \)
        \Comment{Append \( y_t \) to multimodal sequence \( S \)}
        \State \( S \gets S + y_t \)
        \If{\(\langle \text{function} \rangle\) \(\langle /\text{function} \rangle\) detected in \( y_t \)}
            \Comment{Extract tool command \( C_t \)}
            \State \( C_t \gets \text{Parse}(\langle \text{function} \rangle, y_t, \langle /\text{function} \rangle) \)
            \Comment{Retrieve tool instruction}
            \State \( I_t = \mathcal{T}(I, C_t) \)
            \Comment{Image operation using the tool box}
            \State \( S \gets S + \langle \text{observation} \rangle \ I_t \langle /\text{observation} \rangle \)
            \State Increment round \( r \gets r + 1 \)
        \EndIf
        \If{\(\langle \text{answer} \rangle\) \(\langle /\text{answer} \rangle\) detected in \( S \)}
            \Comment{Terminate generation}
            \State \Return final generated response \( S \)
        \EndIf
    \EndWhile
    \State \Return final generated response \( S \)
\end{algorithmic}
\end{algorithm}

\section{Experiments}

\subsection{Experimental Setup}

\subsubsection{Implementation Details}
We perform full parameter fine-tuning on Qwen2.5-VL-7B~\cite{bai2025qwen2} using TWI-Tools-146K data, with a mixture of partial general reasoning data. The maximum sequence length is set to 8,192 tokens. Input images are resized to a resolution range between a minimum of $4 \times 28 \times 28$ pixels and a maximum of $1024 \times 28 \times 28$ pixels. We employ a cosine learning rate scheduler with an initial rate of 1.0e-5 and 10\% warmup ratio, training for 1 epoch with the batch size of 8. The vision encoder and multi-modal projector remain frozen during training.

\subsubsection{Baselines and Benchmarks}
We compare Simple o3 with proprietary models like GPT-4o~\cite{hurst2024gpt} and 4o-mini, as well as competitive open-source models, LLaVA-OneVision-7B~\cite{li2024llava} and Ovis1.6-Gemma2-9B~\cite{lu2024ovis}. We additionally evaluate it against other models adopting the ``thinking with images'' paradigm, including DeepEyes~\cite{zheng2025deepeyes} and Chain-of-Focus~\cite{zhang2025chain}. To validate the effectiveness of Simple o3, we evaluate it on multimodal reasoning benchmarks, including the reasoning subtasks in MME~\cite{fu2024mmecomprehensiveevaluationbenchmark} and reasoning questions in CharXiv~\cite{wang2024CharXiv}, VisuLogic~\cite{xu2025visulogic}, as well as fine-grained perception benchmarks such as HR-Bench 4K~\cite{wang2025divide}, VStarBench~\cite{wu2024v}, and COCO Caption~\cite{chen2015microsoft}. Additionally, we assess its performance on general VQA tasks like ScienceQA~\cite{saikh2022scienceqa}, MMStar~\cite{chen2024we}, and MMVet~\cite{yu2024mmvetevaluatinglargemultimodal}. Beyond these comprehensive evaluations, we also test the model on hallucination benchmarks, including POPE~\cite{li2023evaluatingobjecthallucinationlarge} and HallusionBench~\cite{guan2024hallusionbench}, and chart comprehension tasks such as DocVQA~\cite{mathew2021docvqa} and InfoVQA~\cite{mathew2022infographicvqa}.

\subsection{Main Experiments}
By delving into the performances in Table~\ref{tab:main_1}, we conclude that Simple o3 achieves significant improvements on multimodal reasoning benchmarks. Notably, on the MME reasoning subset, Simple o3 outperforms the base model by nearly 50 points—a margin that even surpasses GPT-4o by 27 points. This strongly demonstrates the effectiveness of ``thinking with images'' where increasing the number of attended visual tokens enhances the model's reasoning capability. Additionally, our method achieves improvements of 4.2\% and 3.3\% on CharXiv and VisuLogic respectively. For cases involving multiple abstract entities in images, the model leverages its robust grounding capability to locate relevant sub-regions, and through multi-step atomic reasoning, it gradually approaches the correct answer.

For multimodal perception, COCO Caption focuses on coarse-grained perception, evaluating a model’s holistic understanding of images. Compared to the base model, Simple o3 improves performance by 3.1 points by leveraging increased reuse of the original image to enhance detailed comprehension during inference. Meanwhile, HR-Bench 4K and VStarBench involve high-resolution images where target objects may occupy only a small fraction of pixels, emphasizing fine-grained perception. Our method achieves superior performance, approximately 7.4\% and 12.9\%, on these fine-grained benchmarks, outperforming existing approaches. Beyond the significant improvements in fine-grained perception, as shown in Table~\ref{tab:spatial}, we also observe that the model exhibits strong advantages in handling spatial reasoning tasks, particularly on subtasks of MMVet and LogicVista, achieving improvements of 16.6\% and 10.3\%. This suggests that our vision-language interleaved reasoning approach enhances the model’s understanding of relationships between different entities or areas. By accurately predicting the positional information of target objects and extracting visual features from relevant regions, the model achieves a unified comprehension of object locations across both language and visual modalities.

We also compare Simple o3 with two RL-based approaches, DeepEyes and Chain-of-Focus, which both employ the 'Crop-then-Resize' strategy in Table~\ref{tab:comparison}. Our method demonstrates superior performance on both HR-Bench 4K and VStarBench, further validating the effectiveness of the ``thinking with images'' in visual reasoning. Although Simple o3 does not achieve the best performance compared to other methods in POPE, it still improves on two hallucination benchmarks, especially by 2.3\% in HallusionBench.

In general VQA benchmarks, as shown in Table~\ref{tab:main_2}, Simple o3 demonstrates consistent superiority over Qwen2.5-VL-7B. In ScienceQA, Simple o3 achieves 90.0\%, outperforming by 1.3\%, which reflects its stronger capability in integrating multidisciplinary knowledge for multimodal QA. The model further exhibits enhanced real-world applicability with a 1.1\% improvement on RealWorldQA, showcasing superior generalization in practical scenarios. Notably, Simple o3 advances multimodal integration with significant gains of 2.6 and 1.5 points on MMStar and MMVet respectively, proving its exceptional effectiveness in fusing visual-language data and handling complex multimodal tasks through refined multidimensional capability alignment. Although the performance of foundation models on the DocVQA benchmark has nearly reached saturation at 94.7\%, Simple o3 still maintains its performance lead through fine-tuning. Meanwhile, its improvement on the InfoVQA benchmark demonstrates significantly enhanced capabilities in parsing key information from charts.

% \usepackage{multirow}
\begin{table*}[t]
\small
\centering
\begin{tabular}{l|ccc|ccc|cc}
\toprule
\multirow{2}{*}{Model} & \multicolumn{3}{c|}{Reasoning} & \multicolumn{3}{c|}{Perception} & \multicolumn{2}{c}{Hallucination} \\
 & MME(R) & CharXiv & VisLog & HR-4K & V* & COCOC & POPE & HalB \\ \hline
GPT-4o* & 674.6 & 29.9 & 25.1 & 46.8 & 45.0 & 23.1 & 84.6 & 44.2 \\
GPT4o-mini* & 564.3 & 32.7 & 25.8 & 48.0 & 50.8 & 17.4 & 83.3 & 39.3 \\ \hline
Ovis1.6-7B* & 547.5 & 36.5 & - & 64.1 & 71.2 & 14.4 & 87.8 & 44.1 \\
LLaVA-OV-9B* & 415.4 & 22.4 & 22.5 & 64.7 & 72.8 & 9.9 & 88.3 & 30.3 \\
Qwen2.5-VL-7B* & 652.5 & 37.6 & 26.0 & 68.8 & 77.5 & 15.0 & 86.0 & 42.1 \\ \hline
Simple o3* & 702.1 & 41.8 & 29.3 & 76.2 & 90.4 & 18.1 & 87.6 & 44.4 \\
$\Delta$ ($vs$ Base) & +49.6 & +4.2 & +3.3 & +7.4 & +12.9 & +3.1 & +1.6 & +2.3 \\ 
\bottomrule
\end{tabular}
\caption{Comparison results on multimodal reasoning benchmarks: reasoning tasks in MME, CharXiv(val), and VisuLogic, perception benchmarks: HR-Bench 4K, VStarBench, and COCO Caption(val), and hallucination benchmarks: POPE and HallusionBench. We use Rouge-L scores for COCO Caption. For HallusionBench, we report the mean accuracy of unique questions and all figures. * denotes the result is reproduced by ourselves. For HR-Bench 4K, VStarBench, and POPE, the highest input resolution is set to $16384 \times 28 \times 28$, and for other benchmarks, it is set to $2048 \times 28 \times 28$.}
\label{tab:main_1}
\end{table*}

% \usepackage{multirow}
% \begin{table}[]
% \resizebox{0.47\textwidth}{!}{
% \begin{tabular}{l|cccc|cc}
% \toprule
% \multirow{2}{*}{Model} & \multicolumn{4}{c|}{General VQA} & \multicolumn{2}{c}{Chart} \\
%  & Sci-QA & RW-QA & MM* & MMVet & DocVQA & InfoVQA \\ \hline
% GPT-4o* & 88.4 & 67.2 & 61.5 & 68.6 & 55.5 & 38.9 \\
% GPT4o-mini* & 85.6 & 66.7 & 54.5 & 69.8 & 78.0 & 57.9 \\ \hline
% Ovis1.6-9B* & 93.3 & 71.0 & 62.6 & 61.9 & 88.9 & 73.4 \\
% LLaVA-OV-7B* & 95.3 & 69.7 & 61.9 & 52.7 & 87.0 & 66.4 \\
% Qwen2.5-VL-7B* & 88.7 & 68.4 & 63.9 & 68.2 & 94.7 & 80.0 \\ \hline
% Simple o3* & 90.0 & 69.5 & 66.5 & 69.7 & 94.8 & 82.0 \\
% $\Delta$ ($vs$ Base) & +1.3 & +1.1 & +2.6 & +1.5 & +0.1 & +2.0 \\ 
% \bottomrule
% \end{tabular}}
% \caption{Comparison results on general VQA benchmarks: ScienceQA, RealWorldQA, MMStar, and MMVet, and chart comprehension benchmarks: DocVQA(val) and InfoVQA(val). The highest input resolution is set to $2048 \times 28 \times 28$.}
% \label{tab:main_2}
% \end{table}

\begin{table}[]
\setlength{\tabcolsep}{1pt}
\small
\begin{tabular}{l|cccc|cc}
\toprule
\multirow{2}{*}{Model} & \multicolumn{4}{c|}{General VQA} & \multicolumn{2}{c}{Chart} \\
 & Sci-QA & RW-QA & MM* & MMVet & DocVQA & InfoVQA \\ \hline
GPT-4o* & 88.4 & 67.2 & 61.5 & 68.6 & 55.5 & 38.9 \\
GPT4o-mini* & 85.6 & 66.7 & 54.5 & 69.8 & 78.0 & 57.9 \\ \hline
Ovis1.6-9B* & 93.3 & 71.0 & 62.6 & 61.9 & 88.9 & 73.4 \\
LLaVA-OV-7B* & 95.3 & 69.7 & 61.9 & 52.7 & 87.0 & 66.4 \\
Qwen2.5-VL-7B* & 88.7 & 68.4 & 63.9 & 68.2 & 94.7 & 80.0 \\ \hline
Simple o3* & 90.0 & 69.5 & 66.5 & 69.7 & 94.8 & 82.0 \\
$\Delta$ ($vs$ Base) & +1.3 & +1.1 & +2.6 & +1.5 & +0.1 & +2.0 \\ 
\bottomrule
\end{tabular}
\caption{Comparison results on general VQA benchmarks: ScienceQA, RealWorldQA, MMStar, and MMVet, and chart comprehension benchmarks: DocVQA(val) and InfoVQA(val). The highest input resolution is set to $2048 \times 28 \times 28$.}
\label{tab:main_2}
\end{table}

% \usepackage{multirow}
\begin{table}[]
\begin{tabular}{l|ccc}
\toprule
\multirow{2}{*}{Model} & VisuLogic & MMVet & LogicVista \\
 & Spatial & rec\_spat & spatial \\ \hline
Qwen2.5-VL-7B* & 22.9 & 66.7 & 20.5 \\ \hline
Simple o3* & 26.4 & 83.3 & 30.8 \\
$\Delta$ ($vs$ Base) & +3.5 & +16.6 & +10.3 \\ \bottomrule
\end{tabular}
\caption{The performance of Simple o3 on spatially relevant benchmarks.}
\label{tab:spatial}
\end{table}

\begin{table}[]
\begin{tabular}{l|ccc}
\toprule
Model & \multicolumn{1}{l}{HR-Bench 4K} & \multicolumn{1}{l}{POPE} & \multicolumn{1}{l}{V*Bench} \\ \hline
Qwen2.5-VL-7B* & 68.8 & 86.0 & 77.5 \\
DeepEyes & 75.1 & 87.7 & 90.1 \\
Chain-of-Focus & - & 88.4 & 88.0 \\ \hline
Simple o3* & 76.2 & 87.6 & 90.4 \\
$\Delta$ ($vs$ Best) & +1.1 & -0.8 & +0.3 \\
\bottomrule
\end{tabular}
\caption{Comparative performance of Simple o3 with other interleaved vision-language reasoning methods: DeepEyes and Chain-of-Focus.}
\label{tab:comparison}
\end{table}

\subsection{Ablation Study}
For the ablation study, we select representative benchmarks from key categories including multimodal reasoning, perception, hallucination, and chart understanding. We analyze the impact of visual manipulation tools in Table~\ref{tab:tool_implementation}, by incrementally incorporating components, beginning with $reuse$, to quantify their cumulative contributions. Experiments demonstrate that the $reuse$ operation delivers significant performance improvements, achieving gains of 31.2 points on MME and 4.8\% on VStarBench, with at least 1-point improvements on other benchmarks. This indicates that $reuse$ enhances the model's perception and reasoning capabilities through the introduction of additional visual tokens. While all operations contribute positively to performance, the $zoom\_in$ operation yields marginal benefits (mostly below 1\%) except on VStarBench. $focus\_area$ emerges as another critical tool, particularly excelling in fine-grained perception tasks like HR-Bench 4K and VStarBench with improvements of 4.9\% and 6.9\% respectively. Its advantage lies in precise visual localization that eliminates redundant visual information while focusing on key entities or regions, thereby delivering consistent performance gains. These findings suggest that the synergistic combination of these three tools elevates the model's reasoning upper bound.

Further, our analysis extends to the role of heterogeneous training data in shaping model capabilities, as summarized in Table~\ref{tab:training_data}. First, general reasoning data results in a slight overall boost in model performance. MATHV360K broadens the data distribution by introducing complex visual QA, charts, logical reasoning problems. This leads to a performance leap of nearly 27 points on the MME reasoning task, alongside gains of 1.3\% on CharXiv and 0.9\% on InfoVQA. We attribute this to the large scale of complex VQA and chart data, which strengthen the model's representation of logical inter-object relations, thereby refining its strategic planning for visual tool operations. Concurrently, the logic-centric tasks directly bolster its atomic reasoning capabilities. However, a relative dearth of perceptual samples in the initial training mix results in modest gains on fine-grained perception benchmarks, including HR-Bench 4K, VStarBench, and HallusionBench. To remediate this deficit, we introduce perceptual data from LLaVA-CoT-100K, an intervention that elevates the model's fine-grained perception capability without a trade-off in reasoning performance, specifically improving by 3.6\% on HR-Bench 4K and 6.8\% on VStarBench. We further expanded the training dataset by incorporating ArxivQA, which consists of challenging abstract diagrams from Arxiv papers, substantially fortifies the model's logical reasoning faculties despite its narrow data distribution, improving it by 13.5 points on the MME reasoning task and up to 2.1\% on CharXiv. Ultimately, the synergistic integration of multi-source, heterogeneous datasets enhances and amplifies the model's higher-order iterative reasoning capabilities.

\begin{table}[]
\small
\setlength{\tabcolsep}{1pt}
% \resizebox{0.47\textwidth}{!}{
\begin{tabular}{l|cccccc}
\toprule
Model & MME(R) & CharXiv & HR-4K & V* & HalB & InfoVQA \\ \hline
Qwen2.5-VL-7B & 652.5 & 37.6 & 68.8 & 77.5 & 42.1 & 80.0 \\ \hline
+ w/ $reuse$ & 683.7 & 39.0 & 70.5 & 82.3 & 43.3 & 81.6 \\
++ w/ $zoom\_in$ & 689.6 & 39.6 & 71.3 & 83.5 & 43.7 & 81.8 \\
+++ w/ $focus\_area$ & \textbf{702.1} & \textbf{41.8} & \textbf{76.2} & \textbf{90.4} & \textbf{44.4} & \textbf{82.0} \\ 
\bottomrule
\end{tabular}
\caption{Performance comparison of tool implementations variants: Only $reuse$, plus $zoom\_in$ , and with $focus\_area$.}
\label{tab:tool_implementation}
\end{table}

\begin{table}[]
\small
\setlength{\tabcolsep}{1pt}
% \resizebox{0.47\textwidth}{!}{
\begin{tabular}{l|cccccc}
\toprule
Model & MME(R) & CharXiv & HR-4K & V* & HalB & InfoVQA \\ \hline
Qwen2.5-VL-7B & 652.5 & 37.6 & 68.8 & 77.5 & 42.1 & 80.0 \\ 
w/ general & 656.8 & 38.1 & 69.2 & 77.6 & 42.3 & 80.3 \\ \hline
+ w/ MATHV360K & 683.4 & 39.4 & 71.3 & 81.4 & 42.8 & 81.2 \\
++ w/ LLaVA-CoT & 688.6 & 39.7 & 74.9 & 88.2 & 44.1 & 81.4 \\
+++ w/ ArxivQA & \textbf{702.1} & \textbf{41.8} & \textbf{76.2} & \textbf{90.4} & \textbf{44.4} & \textbf{82.0} \\ 
\bottomrule
\end{tabular}
\caption{Impact of training data on model performance: Only MATHV360K, plus LLaVA-CoT-100K, and with ArxivQA.}
\label{tab:training_data}
\end{table}

\subsection{Further Analysis}
\subsubsection{Should We Crop, Draw Bbox, or Reuse the Original Images When Focusing on Key Entities or Regions?} 

During training, we draw bbox on images based on the coordinates returned by the $focus\_area$ operation. This stems from the fact that some coordinates only provide approximate focal regions, analogous to human visual attention mechanisms, which may either incompletely cover target instances or excessively include irrelevant areas. Directly cropping images based on samples with incomplete target coverage would lead to critical visual information loss, thereby compromising the reasoning chain integrity. 

To identify the optimal image manipulation method when focusing on key entities or regions, we evaluate three $focus\_area$ modes—drawing the bbox, image cropping, and reusing the original image—while preserving the functionalities of $zoom\_in$ and $reuse$. As shown in Table~\ref{tab:focus_area}, experimental results demonstrate that the image cropping delivers optimal performance. Specifically, Simple o3 can accurately cover target entities or regions when executing $focus\_area$ operations, without producing incomplete crops that trigger ``I can't see'' or ``I can't find'' responses. This proves that repeatedly injecting key visual information effectively enhances model reasoning capabilities.

Notably, although the input distribution of cropped images (which may contain only partial entities) differs from the distribution of training data (mostly complete multi-entity images), this discrepancy does not affect model performance. Benefiting from the strong grounding capability of the base model, Simple o3 further unleashes its reasoning potential while learning this interleaved image-text reasoning format. In contrast, while the method of drawing bbox aligns with the training distribution and preserves complete entity information, those redundant visual tokens actually degrade model performance. Reusing the original image yields the poorest results due to the lack of explicit region indication, which tends to scatter attention across irrelevant entities or regions.

\begin{table}
\small
\setlength{\tabcolsep}{3.5pt}
% \resizebox{0.47\textwidth}{!}{
\begin{tabular}{l|cccccc}
\toprule
Method & MME(R) & CharXiv & HR-4K & V* & HalB & InfoVQA \\ \hline
Crop & 702.1 & \textbf{41.8} & \textbf{76.2} & 90.4 & \textbf{44.4} & \textbf{82.0} \\
Draw & 698.2 & 41.2 & 75.4 & \textbf{90.6} & 43.8 & \textbf{82.0} \\
Reuse & \textbf{704.3} & 40.7 & 74.2 & 89.3 & 43.8 & 81.8 \\ \bottomrule
\end{tabular}
\caption{Performance comparison of $focus\_area$ operation variants: image cropping, bbox drawing, and original image reusing, based on provided coordinates.}
\label{tab:focus_area}
\end{table}

\subsubsection{How Does Input Resolution Affect Model Performance?}
We explored the impact of the maximum input image resolution on the performance of Simple o3, dividing it into three variants: low resolution, medium resolution, and high resolution. As shown in Table~\ref{tab:resolution}, When the maximum input resolution was set to a low level, the model performed poorest—an expected outcome, since excessively low-resolution images impair the model's fundamental perceptual capabilities. Despite being trained primarily on low-resolution images, the model generalizes well to higher-resolution inputs during inference. On multimodal reasoning benchmarks, medium resolution proved sufficient for capturing core visual elements, and increasing the input from medium to high resolution did not yield significant performance gains. Notably, in the CharXiv task, medium-resolution input even outperformed high-resolution input by 1.7\%. This may due to the explosion of visual tokens at high resolutions, which could disrupt the model's attention allocation across long sequences and hinder effective focus on critical regions. However, high-resolution inputs demonstrably enhanced performance on perception-oriented benchmarks while further reducing hallucination phenomena and improving parsing accuracy in document-based tasks.

\begin{table}[]
\small
\setlength{\tabcolsep}{4pt}
% \resizebox{0.47\textwidth}{!}{
\begin{tabular}{l|cccccc}
\toprule
Res & MME(R) & CharXiv & HR-4K & V* & HalB & InfoVQA \\ \hline
Low & 699.2 & 32.4 & 56.9 & 61.3 & 41.5 & 62.3 \\
Med & \textbf{702.1} & \textbf{41.8} & 69.3 & 78.0 & 44.4 & 82.0 \\
High & \textbf{702.1} & 40.1 & \textbf{76.2} & \textbf{90.4} & \textbf{46.7} & \textbf{82.6} \\ \bottomrule
\end{tabular}
\caption{Performance comparison of maximum input resolution variants: low resolution ($256 \times 28 \times 28$), medium resolution ($2048 \times 28 \times 28$), and high resolution ($16384 \times 28 \times 28$).}
\label{tab:resolution}
\end{table}

\subsection{Conclusion}
In this work, we present Simple o3, a novel framework that significantly advances multimodal reasoning capabilities through dynamic tool interactions and interleaved vision-language reasoning. By introducing a scalable, low-cost data synthesis pipeline, we curate the high-quality TWI-Tools-146K dataset, establishing a foundation for training models on complex reasoning tasks. Incorporating image masking during training, Simple o3 demonstrates strong performance on diverse benchmarks. Furthermore, we present the first in-depth analysis of how different tools affect model inference performance, providing key insights for advancing the 'thinking with images' paradigm. We found that by introducing additional visual tokens, reusing and magnifying the original image improves the model's visual reasoning and fine-grained perception, while image cropping based on precise visual grounding allows the model to effectively focus on key entities or regions, further enhancing its capabilities. In the future, we will explore more tools and thinking with images with RL training, and extend the paradigm from vision-language reasoning model to vision-language action model.

\bibliography{main}

\begin{thebibliography}{62}
\providecommand{\natexlab}[1]{#1}

\bibitem[{Alayrac et~al.(2022)Alayrac, Donahue, Luc, Miech, Barr, Hasson, Lenc, Mensch, Millican, Reynolds et~al.}]{alayrac2022flamingo}
Alayrac, J.-B.; Donahue, J.; Luc, P.; Miech, A.; Barr, I.; Hasson, Y.; Lenc, K.; Mensch, A.; Millican, K.; Reynolds, M.; et~al. 2022.
\newblock Flamingo: a visual language model for few-shot learning.
\newblock \emph{Advances in neural information processing systems}, 35: 23716--23736.

\bibitem[{Awadalla et~al.(2023)Awadalla, Gao, Gardner, Hessel, Hanafy, Zhu, Marathe, Bitton, Gadre, Sagawa et~al.}]{awadalla2023openflamingo}
Awadalla, A.; Gao, I.; Gardner, J.; Hessel, J.; Hanafy, Y.; Zhu, W.; Marathe, K.; Bitton, Y.; Gadre, S.; Sagawa, S.; et~al. 2023.
\newblock Openflamingo: An open-source framework for training large autoregressive vision-language models.
\newblock \emph{arXiv preprint arXiv:2308.01390}.

\bibitem[{Bai et~al.(2023)Bai, Bai, Chu, Cui, Dang, Deng, Fan, Ge, Han, Huang et~al.}]{bai2023qwen}
Bai, J.; Bai, S.; Chu, Y.; Cui, Z.; Dang, K.; Deng, X.; Fan, Y.; Ge, W.; Han, Y.; Huang, F.; et~al. 2023.
\newblock Qwen technical report.
\newblock \emph{arXiv preprint arXiv:2309.16609}.

\bibitem[{Bai et~al.(2025{\natexlab{a}})Bai, Chen, Liu, Wang, Ge, Song, Dang, Wang, Wang, Tang et~al.}]{bai2025qwen2}
Bai, S.; Chen, K.; Liu, X.; Wang, J.; Ge, W.; Song, S.; Dang, K.; Wang, P.; Wang, S.; Tang, J.; et~al. 2025{\natexlab{a}}.
\newblock Qwen2. 5-vl technical report.
\newblock \emph{arXiv preprint arXiv:2502.13923}.

\bibitem[{Bai et~al.(2025{\natexlab{b}})Bai, Hu, Sun, Qiu, Jiang, He, Zeng, He, Yuan, and Zhang}]{bai2025multi}
Bai, T.; Hu, Z.; Sun, F.; Qiu, J.; Jiang, Y.; He, G.; Zeng, B.; He, C.; Yuan, B.; and Zhang, W. 2025{\natexlab{b}}.
\newblock Multi-Step Visual Reasoning with Visual Tokens Scaling and Verification.
\newblock \emph{arXiv preprint arXiv:2506.07235}.

\bibitem[{Chen et~al.(2023)Chen, Zhang, Zeng, Zhang, Zhu, and Zhao}]{chen2023shikra}
Chen, K.; Zhang, Z.; Zeng, W.; Zhang, R.; Zhu, F.; and Zhao, R. 2023.
\newblock Shikra: Unleashing multimodal llm's referential dialogue magic.
\newblock \emph{arXiv preprint arXiv:2306.15195}.

\bibitem[{Chen et~al.(2024{\natexlab{a}})Chen, Li, Dong, Zhang, Zang, Chen, Duan, Wang, Qiao, Lin et~al.}]{chen2024we}
Chen, L.; Li, J.; Dong, X.; Zhang, P.; Zang, Y.; Chen, Z.; Duan, H.; Wang, J.; Qiao, Y.; Lin, D.; et~al. 2024{\natexlab{a}}.
\newblock Are we on the right way for evaluating large vision-language models?
\newblock \emph{Advances in Neural Information Processing Systems}, 37: 27056--27087.

\bibitem[{Chen et~al.(2015)Chen, Fang, Lin, Vedantam, Gupta, Doll{\'a}r, and Zitnick}]{chen2015microsoft}
Chen, X.; Fang, H.; Lin, T.-Y.; Vedantam, R.; Gupta, S.; Doll{\'a}r, P.; and Zitnick, C.~L. 2015.
\newblock Microsoft coco captions: Data collection and evaluation server.
\newblock \emph{arXiv preprint arXiv:1504.00325}.

\bibitem[{Chen et~al.(2024{\natexlab{b}})Chen, Wang, Cao, Liu, Gao, Cui, Zhu, Ye, Tian, Liu et~al.}]{chen2024expanding}
Chen, Z.; Wang, W.; Cao, Y.; Liu, Y.; Gao, Z.; Cui, E.; Zhu, J.; Ye, S.; Tian, H.; Liu, Z.; et~al. 2024{\natexlab{b}}.
\newblock Expanding performance boundaries of open-source multimodal models with model, data, and test-time scaling.
\newblock \emph{arXiv preprint arXiv:2412.05271}.

\bibitem[{Chung et~al.(2025)Chung, Kim, Kim, Lee, Kim, and Yu}]{chung2025don}
Chung, J.; Kim, J.; Kim, S.; Lee, J.; Kim, M.~S.; and Yu, Y. 2025.
\newblock Don't Look Only Once: Towards Multimodal Interactive Reasoning with Selective Visual Revisitation.
\newblock \emph{arXiv preprint arXiv:2505.18842}.

\bibitem[{Dai et~al.(2023)Dai, Li, Li, Tiong, Zhao, Wang, Li, Fung, and Hoi}]{dai2023instructblip}
Dai, W.; Li, J.; Li, D.; Tiong, A.; Zhao, J.; Wang, W.; Li, B.; Fung, P.~N.; and Hoi, S. 2023.
\newblock Instructblip: Towards general-purpose vision-language models with instruction tuning.
\newblock \emph{Advances in neural information processing systems}, 36: 49250--49267.

\bibitem[{Fan et~al.(2024)Fan, Ji, Jiang, Li, Jin, Song, Wang, Hong, Chen, Zheng et~al.}]{fan2024mousi}
Fan, X.; Ji, T.; Jiang, C.; Li, S.; Jin, S.; Song, S.; Wang, J.; Hong, B.; Chen, L.; Zheng, G.; et~al. 2024.
\newblock Mousi: Poly-visual-expert vision-language models.
\newblock \emph{arXiv preprint arXiv:2401.17221}.

\bibitem[{Fan et~al.(2025)Fan, He, Yang, Zheng, Kuo, Zheng, Narayanaraju, Guan, and Wang}]{fan2025gritteachingmllmsthink}
Fan, Y.; He, X.; Yang, D.; Zheng, K.; Kuo, C.-C.; Zheng, Y.; Narayanaraju, S.~J.; Guan, X.; and Wang, X.~E. 2025.
\newblock GRIT: Teaching MLLMs to Think with Images.
\newblock arXiv:2505.15879.

\bibitem[{Fu et~al.(2024)Fu, Chen, Shen, Qin, Zhang, Lin, Yang, Zheng, Li, Sun, Wu, and Ji}]{fu2024mmecomprehensiveevaluationbenchmark}
Fu, C.; Chen, P.; Shen, Y.; Qin, Y.; Zhang, M.; Lin, X.; Yang, J.; Zheng, X.; Li, K.; Sun, X.; Wu, Y.; and Ji, R. 2024.
\newblock MME: A Comprehensive Evaluation Benchmark for Multimodal Large Language Models.
\newblock arXiv:2306.13394.

\bibitem[{Ge et~al.(2024)Ge, Cheng, Wang, Yuan, Gao, Song, Song, Huang, and Zheng}]{ge2024convllava}
Ge, C.; Cheng, S.; Wang, Z.; Yuan, J.; Gao, Y.; Song, J.; Song, S.; Huang, G.; and Zheng, B. 2024.
\newblock Convllava: Hierarchical backbones as visual encoder for large multimodal models.
\newblock \emph{arXiv preprint arXiv:2405.15738}.

\bibitem[{Guan et~al.(2024)Guan, Liu, Wu, Xian, Li, Liu, Wang, Chen, Huang, Yacoob et~al.}]{guan2024hallusionbench}
Guan, T.; Liu, F.; Wu, X.; Xian, R.; Li, Z.; Liu, X.; Wang, X.; Chen, L.; Huang, F.; Yacoob, Y.; et~al. 2024.
\newblock Hallusionbench: an advanced diagnostic suite for entangled language hallucination and visual illusion in large vision-language models.
\newblock In \emph{Proceedings of the IEEE/CVF Conference on Computer Vision and Pattern Recognition}, 14375--14385.

\bibitem[{Guo et~al.(2025)Guo, Yang, Zhang, Song, Zhang, Xu, Zhu, Ma, Wang, Bi et~al.}]{guo2025deepseek}
Guo, D.; Yang, D.; Zhang, H.; Song, J.; Zhang, R.; Xu, R.; Zhu, Q.; Ma, S.; Wang, P.; Bi, X.; et~al. 2025.
\newblock Deepseek-r1: Incentivizing reasoning capability in llms via reinforcement learning.
\newblock \emph{arXiv preprint arXiv:2501.12948}.

\bibitem[{Huang et~al.(2025)Huang, Jia, Zhai, Cao, Ye, Zhao, Xu, Hu, and Lin}]{huang2025vision}
Huang, W.; Jia, B.; Zhai, Z.; Cao, S.; Ye, Z.; Zhao, F.; Xu, Z.; Hu, Y.; and Lin, S. 2025.
\newblock Vision-r1: Incentivizing reasoning capability in multimodal large language models.
\newblock \emph{arXiv preprint arXiv:2503.06749}.

\bibitem[{Hurst et~al.(2024)Hurst, Lerer, Goucher, Perelman, Ramesh, Clark, Ostrow, Welihinda, Hayes, Radford et~al.}]{hurst2024gpt}
Hurst, A.; Lerer, A.; Goucher, A.~P.; Perelman, A.; Ramesh, A.; Clark, A.; Ostrow, A.; Welihinda, A.; Hayes, A.; Radford, A.; et~al. 2024.
\newblock Gpt-4o system card.
\newblock \emph{arXiv preprint arXiv:2410.21276}.

\bibitem[{Jin et~al.(2025)Jin, Zeng, Yue, Yoon, Arik, Wang, Zamani, and Han}]{jin2025search}
Jin, B.; Zeng, H.; Yue, Z.; Yoon, J.; Arik, S.; Wang, D.; Zamani, H.; and Han, J. 2025.
\newblock Search-r1: Training llms to reason and leverage search engines with reinforcement learning.
\newblock \emph{arXiv preprint arXiv:2503.09516}.

\bibitem[{Kojima et~al.(2022)Kojima, Gu, Reid, Matsuo, and Iwasawa}]{kojima2022large}
Kojima, T.; Gu, S.~S.; Reid, M.; Matsuo, Y.; and Iwasawa, Y. 2022.
\newblock Large language models are zero-shot reasoners.
\newblock \emph{Advances in neural information processing systems}, 35: 22199--22213.

\bibitem[{Li et~al.(2024{\natexlab{a}})Li, Zhang, Guo, Zhang, Li, Zhang, Zhang, Zhang, Li, Liu et~al.}]{li2024llava}
Li, B.; Zhang, Y.; Guo, D.; Zhang, R.; Li, F.; Zhang, H.; Zhang, K.; Zhang, P.; Li, Y.; Liu, Z.; et~al. 2024{\natexlab{a}}.
\newblock Llava-onevision: Easy visual task transfer.
\newblock \emph{arXiv preprint arXiv:2408.03326}.

\bibitem[{Li et~al.(2023{\natexlab{a}})Li, Li, Savarese, and Hoi}]{li2023blip}
Li, J.; Li, D.; Savarese, S.; and Hoi, S. 2023{\natexlab{a}}.
\newblock Blip-2: Bootstrapping language-image pre-training with frozen image encoders and large language models.
\newblock In \emph{International conference on machine learning}, 19730--19742. PMLR.

\bibitem[{Li et~al.(2024{\natexlab{b}})Li, Wang, Xu, Wang, Feng, Kong, and Liu}]{li2024multimodal}
Li, L.; Wang, Y.; Xu, R.; Wang, P.; Feng, X.; Kong, L.; and Liu, Q. 2024{\natexlab{b}}.
\newblock Multimodal arxiv: A dataset for improving scientific comprehension of large vision-language models.
\newblock \emph{arXiv preprint arXiv:2403.00231}.

\bibitem[{Li, Zou, and Liu(2025)}]{li2025torl}
Li, X.; Zou, H.; and Liu, P. 2025.
\newblock Torl: Scaling tool-integrated rl.
\newblock \emph{arXiv preprint arXiv:2503.23383}.

\bibitem[{Li et~al.(2023{\natexlab{b}})Li, Du, Zhou, Wang, Zhao, and Wen}]{li2023evaluatingobjecthallucinationlarge}
Li, Y.; Du, Y.; Zhou, K.; Wang, J.; Zhao, W.~X.; and Wen, J.-R. 2023{\natexlab{b}}.
\newblock Evaluating Object Hallucination in Large Vision-Language Models.
\newblock arXiv:2305.10355.

\bibitem[{Li et~al.(2024{\natexlab{c}})Li, Yang, Liu, Ma, Zhang, Yang, Sun, Liu, and Bai}]{li2024monkey}
Li, Z.; Yang, B.; Liu, Q.; Ma, Z.; Zhang, S.; Yang, J.; Sun, Y.; Liu, Y.; and Bai, X. 2024{\natexlab{c}}.
\newblock Monkey: Image resolution and text label are important things for large multi-modal models.
\newblock In \emph{proceedings of the IEEE/CVF conference on computer vision and pattern recognition}, 26763--26773.

\bibitem[{Liu et~al.(2024{\natexlab{a}})Liu, Li, Li, and Lee}]{liu2024improved}
Liu, H.; Li, C.; Li, Y.; and Lee, Y.~J. 2024{\natexlab{a}}.
\newblock Improved baselines with visual instruction tuning.
\newblock In \emph{Proceedings of the IEEE/CVF conference on computer vision and pattern recognition}, 26296--26306.

\bibitem[{Liu et~al.(2023)Liu, Li, Wu, and Lee}]{liu2023visual}
Liu, H.; Li, C.; Wu, Q.; and Lee, Y.~J. 2023.
\newblock Visual instruction tuning.
\newblock \emph{Advances in neural information processing systems}, 36: 34892--34916.

\bibitem[{Liu et~al.(2024{\natexlab{b}})Liu, Li, Wu, and Lee}]{liu2024visual}
Liu, H.; Li, C.; Wu, Q.; and Lee, Y.~J. 2024{\natexlab{b}}.
\newblock Visual instruction tuning.
\newblock \emph{Advances in neural information processing systems}, 36.

\bibitem[{Liu et~al.(2025{\natexlab{a}})Liu, Peng, Zhong, Yue, Lu, Yu, and Jia}]{liu2025seg}
Liu, Y.; Peng, B.; Zhong, Z.; Yue, Z.; Lu, F.; Yu, B.; and Jia, J. 2025{\natexlab{a}}.
\newblock Seg-zero: Reasoning-chain guided segmentation via cognitive reinforcement.
\newblock \emph{arXiv preprint arXiv:2503.06520}.

\bibitem[{Liu et~al.(2025{\natexlab{b}})Liu, Qu, Zhong, Peng, Liu, Yu, and Jia}]{liu2025visionreasoner}
Liu, Y.; Qu, T.; Zhong, Z.; Peng, B.; Liu, S.; Yu, B.; and Jia, J. 2025{\natexlab{b}}.
\newblock VisionReasoner: Unified Visual Perception and Reasoning via Reinforcement Learning.
\newblock \emph{arXiv preprint arXiv:2505.12081}.

\bibitem[{Liu et~al.(2025{\natexlab{c}})Liu, Sun, Zang, Dong, Cao, Duan, Lin, and Wang}]{liu2025visual}
Liu, Z.; Sun, Z.; Zang, Y.; Dong, X.; Cao, Y.; Duan, H.; Lin, D.; and Wang, J. 2025{\natexlab{c}}.
\newblock Visual-rft: Visual reinforcement fine-tuning.
\newblock \emph{arXiv preprint arXiv:2503.01785}.

\bibitem[{Lu et~al.(2024)Lu, Li, Chen, Xu, Luo, Zhang, and Ye}]{lu2024ovis}
Lu, S.; Li, Y.; Chen, Q.-G.; Xu, Z.; Luo, W.; Zhang, K.; and Ye, H.-J. 2024.
\newblock Ovis: Structural embedding alignment for multimodal large language model.
\newblock \emph{arXiv preprint arXiv:2405.20797}.

\bibitem[{Mathew et~al.(2022)Mathew, Bagal, Tito, Karatzas, Valveny, and Jawahar}]{mathew2022infographicvqa}
Mathew, M.; Bagal, V.; Tito, R.; Karatzas, D.; Valveny, E.; and Jawahar, C. 2022.
\newblock Infographicvqa.
\newblock In \emph{Proceedings of the IEEE/CVF Winter Conference on Applications of Computer Vision}, 1697--1706.

\bibitem[{Mathew, Karatzas, and Jawahar(2021)}]{mathew2021docvqa}
Mathew, M.; Karatzas, D.; and Jawahar, C. 2021.
\newblock Docvqa: A dataset for vqa on document images.
\newblock In \emph{Proceedings of the IEEE/CVF winter conference on applications of computer vision}, 2200--2209.

\bibitem[{Ni et~al.(2025)Ni, Yang, Li, Lin, Lin, Zuo, and Wang}]{ni2025point}
Ni, M.; Yang, Z.; Li, L.; Lin, C.-C.; Lin, K.; Zuo, W.; and Wang, L. 2025.
\newblock Point-rft: Improving multimodal reasoning with visually grounded reinforcement finetuning.
\newblock \emph{arXiv preprint arXiv:2505.19702}.

\bibitem[{Saikh et~al.(2022)Saikh, Ghosal, Mittal, Ekbal, and Bhattacharyya}]{saikh2022scienceqa}
Saikh, T.; Ghosal, T.; Mittal, A.; Ekbal, A.; and Bhattacharyya, P. 2022.
\newblock Scienceqa: A novel resource for question answering on scholarly articles.
\newblock \emph{International Journal on Digital Libraries}, 23(3): 289--301.

\bibitem[{Shao et~al.(2024)Shao, Qian, Xiao, Song, Zong, Wang, Liu, and Li}]{shao2024visual}
Shao, H.; Qian, S.; Xiao, H.; Song, G.; Zong, Z.; Wang, L.; Liu, Y.; and Li, H. 2024.
\newblock Visual cot: Unleashing chain-of-thought reasoning in multi-modal language models.
\newblock \emph{CoRR}.

\bibitem[{Shi et~al.(2024)Shi, Hu, Bin, Liu, Yang, Ng, Bing, and Lee}]{shi2024math}
Shi, W.; Hu, Z.; Bin, Y.; Liu, J.; Yang, Y.; Ng, S.-K.; Bing, L.; and Lee, R. K.-W. 2024.
\newblock Math-llava: Bootstrapping mathematical reasoning for multimodal large language models.
\newblock \emph{arXiv preprint arXiv:2406.17294}.

\bibitem[{Su et~al.(2025)Su, Wang, Ren, Lin, and Chen}]{su2025pixel}
Su, A.; Wang, H.; Ren, W.; Lin, F.; and Chen, W. 2025.
\newblock Pixel reasoner: Incentivizing pixel-space reasoning with curiosity-driven reinforcement learning.
\newblock \emph{arXiv preprint arXiv:2505.15966}.

\bibitem[{Team et~al.(2025)Team, Du, Gao, Xing, Jiang, Chen, Li, Xiao, Du, Liao et~al.}]{team2025kimi}
Team, K.; Du, A.; Gao, B.; Xing, B.; Jiang, C.; Chen, C.; Li, C.; Xiao, C.; Du, C.; Liao, C.; et~al. 2025.
\newblock Kimi k1. 5: Scaling reinforcement learning with llms.
\newblock \emph{arXiv preprint arXiv:2501.12599}.

\bibitem[{Wang et~al.(2025{\natexlab{a}})Wang, Ding, Zeng, Chen, Chen, Wang, Xie, Huang, and Zhao}]{wang2025vrag}
Wang, Q.; Ding, R.; Zeng, Y.; Chen, Z.; Chen, L.; Wang, S.; Xie, P.; Huang, F.; and Zhao, F. 2025{\natexlab{a}}.
\newblock VRAG-RL: Empower Vision-Perception-Based RAG for Visually Rich Information Understanding via Iterative Reasoning with Reinforcement Learning.
\newblock \emph{arXiv preprint arXiv:2505.22019}.

\bibitem[{Wang et~al.(2025{\natexlab{b}})Wang, Ding, Zeng, Zhou, Shen, Luo, Yu, and Tao}]{wang2025divide}
Wang, W.; Ding, L.; Zeng, M.; Zhou, X.; Shen, L.; Luo, Y.; Yu, W.; and Tao, D. 2025{\natexlab{b}}.
\newblock Divide, conquer and combine: A training-free framework for high-resolution image perception in multimodal large language models.
\newblock In \emph{Proceedings of the AAAI Conference on Artificial Intelligence}, volume~39, 7907--7915.

\bibitem[{Wang et~al.(2024)Wang, Xia, He, Chen, Liu, Zhu, Liang, Wu, Liu, Malladi et~al.}]{wang2024CharXiv}
Wang, Z.; Xia, M.; He, L.; Chen, H.; Liu, Y.; Zhu, R.; Liang, K.; Wu, X.; Liu, H.; Malladi, S.; et~al. 2024.
\newblock Charxiv: Charting gaps in realistic chart understanding in multimodal llms.
\newblock \emph{Advances in Neural Information Processing Systems}, 37: 113569--113697.

\bibitem[{Wei et~al.(2022)Wei, Wang, Schuurmans, Bosma, Xia, Chi, Le, Zhou et~al.}]{wei2022chain}
Wei, J.; Wang, X.; Schuurmans, D.; Bosma, M.; Xia, F.; Chi, E.; Le, Q.~V.; Zhou, D.; et~al. 2022.
\newblock Chain-of-thought prompting elicits reasoning in large language models.
\newblock \emph{Advances in neural information processing systems}, 35: 24824--24837.

\bibitem[{Wu and Xie(2024)}]{wu2024v}
Wu, P.; and Xie, S. 2024.
\newblock V?: Guided visual search as a core mechanism in multimodal llms.
\newblock In \emph{Proceedings of the IEEE/CVF Conference on Computer Vision and Pattern Recognition}, 13084--13094.

\bibitem[{Xu et~al.(2024)Xu, Jin, Li, Song, Sun, and Yuan}]{xu2024llava}
Xu, G.; Jin, P.; Li, H.; Song, Y.; Sun, L.; and Yuan, L. 2024.
\newblock Llava-cot: Let vision language models reason step-by-step.
\newblock \emph{arXiv preprint arXiv:2411.10440}.

\bibitem[{Xu et~al.(2025)Xu, Wang, Wang, Chen, Zhou, Yang, Lu, Li, Wang, Zhu et~al.}]{xu2025visulogic}
Xu, W.; Wang, J.; Wang, W.; Chen, Z.; Zhou, W.; Yang, A.; Lu, L.; Li, H.; Wang, X.; Zhu, X.; et~al. 2025.
\newblock Visulogic: A benchmark for evaluating visual reasoning in multi-modal large language models.
\newblock \emph{arXiv preprint arXiv:2504.15279}.

\bibitem[{Ye et~al.(2023{\natexlab{a}})Ye, Hu, Xu, Ye, Yan, Xu, Li, Tian, Qian, Zhang et~al.}]{ye2023ureader}
Ye, J.; Hu, A.; Xu, H.; Ye, Q.; Yan, M.; Xu, G.; Li, C.; Tian, J.; Qian, Q.; Zhang, J.; et~al. 2023{\natexlab{a}}.
\newblock Ureader: Universal ocr-free visually-situated language understanding with multimodal large language model.
\newblock \emph{arXiv preprint arXiv:2310.05126}.

\bibitem[{Ye et~al.(2023{\natexlab{b}})Ye, Xu, Xu, Ye, Yan, Zhou, Wang, Hu, Shi, Shi et~al.}]{ye2023mplug}
Ye, Q.; Xu, H.; Xu, G.; Ye, J.; Yan, M.; Zhou, Y.; Wang, J.; Hu, A.; Shi, P.; Shi, Y.; et~al. 2023{\natexlab{b}}.
\newblock mplug-owl: Modularization empowers large language models with multimodality.
\newblock \emph{arXiv preprint arXiv:2304.14178}.

\bibitem[{Yu et~al.(2025)Yu, Zhang, Zhu, Yuan, Zuo, Yue, Dai, Fan, Liu, Liu et~al.}]{yu2025dapo}
Yu, Q.; Zhang, Z.; Zhu, R.; Yuan, Y.; Zuo, X.; Yue, Y.; Dai, W.; Fan, T.; Liu, G.; Liu, L.; et~al. 2025.
\newblock Dapo: An open-source llm reinforcement learning system at scale.
\newblock \emph{arXiv preprint arXiv:2503.14476}.

\bibitem[{Yu et~al.(2024)Yu, Yang, Li, Wang, Lin, Liu, Wang, and Wang}]{yu2024mmvetevaluatinglargemultimodal}
Yu, W.; Yang, Z.; Li, L.; Wang, J.; Lin, K.; Liu, Z.; Wang, X.; and Wang, L. 2024.
\newblock MM-Vet: Evaluating Large Multimodal Models for Integrated Capabilities.
\newblock arXiv:2308.02490.

\bibitem[{Yuan et~al.(2024)Yuan, Li, Liu, Tang, Luo, Qin, Zhang, and Zhu}]{yuan2024osprey}
Yuan, Y.; Li, W.; Liu, J.; Tang, D.; Luo, X.; Qin, C.; Zhang, L.; and Zhu, J. 2024.
\newblock Osprey: Pixel understanding with visual instruction tuning.
\newblock In \emph{Proceedings of the IEEE/CVF Conference on Computer Vision and Pattern Recognition}, 28202--28211.

\bibitem[{Zeng et~al.(2024)Zeng, Zhang, Zheng, Xia, Wei, Wei, Zhang, Kong, and Song}]{zeng2024matters}
Zeng, Y.; Zhang, H.; Zheng, J.; Xia, J.; Wei, G.; Wei, Y.; Zhang, Y.; Kong, T.; and Song, R. 2024.
\newblock What matters in training a gpt4-style language model with multimodal inputs?
\newblock In \emph{Proceedings of the 2024 Conference of the North American Chapter of the Association for Computational Linguistics: Human Language Technologies (Volume 1: Long Papers)}, 7930--7957.

\bibitem[{Zhang et~al.(2025{\natexlab{a}})Zhang, Zhong, Xia, Yu, Li, He, Shu, Liu, She, Wang et~al.}]{zhang2025cmmcot}
Zhang, G.; Zhong, T.; Xia, Y.; Yu, Z.; Li, H.; He, W.; Shu, F.; Liu, M.; She, D.; Wang, Y.; et~al. 2025{\natexlab{a}}.
\newblock Cmmcot: Enhancing complex multi-image comprehension via multi-modal chain-of-thought and memory augmentation.
\newblock \emph{arXiv preprint arXiv:2503.05255}.

\bibitem[{Zhang et~al.(2025{\natexlab{b}})Zhang, Huang, Yao, Liu, Zhang, Lu, and Tao}]{zhang2025r1}
Zhang, J.; Huang, J.; Yao, H.; Liu, S.; Zhang, X.; Lu, S.; and Tao, D. 2025{\natexlab{b}}.
\newblock R1-vl: Learning to reason with multimodal large language models via step-wise group relative policy optimization.
\newblock \emph{arXiv preprint arXiv:2503.12937}.

\bibitem[{Zhang et~al.(2025{\natexlab{c}})Zhang, Gao, Zhang, Li, Zhang, Liu, Yuan, Wu, Jia, Zhu et~al.}]{zhang2025chain}
Zhang, X.; Gao, Z.; Zhang, B.; Li, P.; Zhang, X.; Liu, Y.; Yuan, T.; Wu, Y.; Jia, Y.; Zhu, S.-C.; et~al. 2025{\natexlab{c}}.
\newblock Chain-of-Focus: Adaptive Visual Search and Zooming for Multimodal Reasoning via RL.
\newblock \emph{arXiv preprint arXiv:2505.15436}.

\bibitem[{Zheng et~al.(2025)Zheng, Yang, Hong, Zhao, Xu, Yang, Shen, and Yu}]{zheng2025deepeyes}
Zheng, Z.; Yang, M.; Hong, J.; Zhao, C.; Xu, G.; Yang, L.; Shen, C.; and Yu, X. 2025.
\newblock DeepEyes: Incentivizing" Thinking with Images" via Reinforcement Learning.
\newblock \emph{arXiv preprint arXiv:2505.14362}.

\bibitem[{Zhou et~al.(2025)Zhou, Li, Wang, Cheng, Zhou, and Hsieh}]{zhou2025r1}
Zhou, H.; Li, X.; Wang, R.; Cheng, M.; Zhou, T.; and Hsieh, C.-J. 2025.
\newblock R1-Zero's" Aha Moment" in Visual Reasoning on a 2B Non-SFT Model.
\newblock \emph{arXiv preprint arXiv:2503.05132}.

\bibitem[{Zhu et~al.(2023)Zhu, Chen, Shen, Li, and Elhoseiny}]{zhu2023minigpt}
Zhu, D.; Chen, J.; Shen, X.; Li, X.; and Elhoseiny, M. 2023.
\newblock Minigpt-4: Enhancing vision-language understanding with advanced large language models.
\newblock \emph{arXiv preprint arXiv:2304.10592}.

\bibitem[{Zhu et~al.(2025)Zhu, Zhong, Zhao, Du, Huang, Liu, Chen, Zou, Chen, Yang et~al.}]{zhu2025active}
Zhu, M.; Zhong, H.; Zhao, C.; Du, Z.; Huang, Z.; Liu, M.; Chen, H.; Zou, C.; Chen, J.; Yang, M.; et~al. 2025.
\newblock Active-O3: Empowering Multimodal Large Language Models with Active Perception via GRPO.
\newblock \emph{arXiv preprint arXiv:2505.21457}.

\end{thebibliography}

% \input{AuthorKit26/ReproducibilityChecklist/LaTeX/ReproducibilityChecklist}

\appendix

\begin{figure*}[t]
    \centering
    \includegraphics[width=1\textwidth]{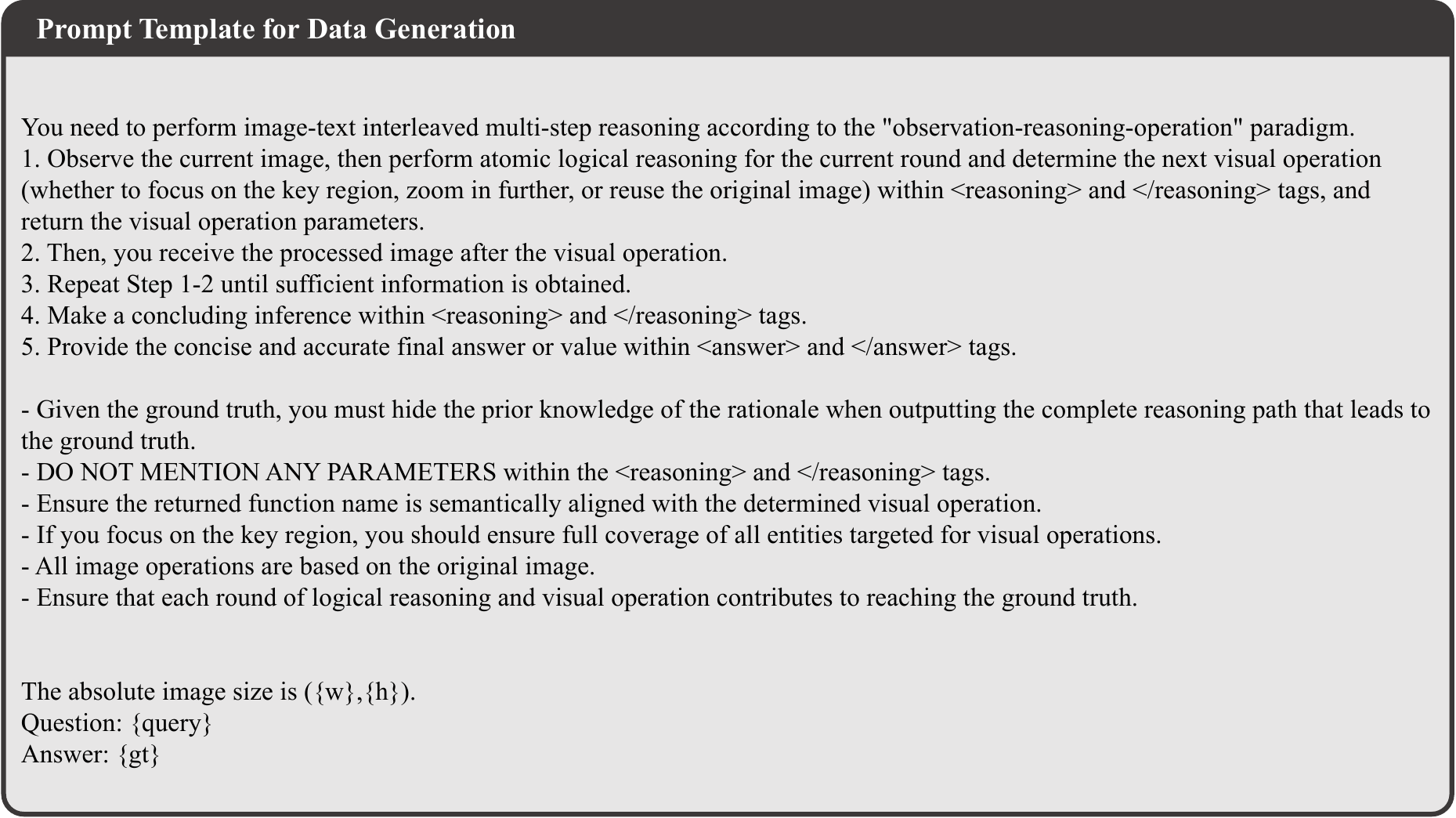}
    \caption{Prompt template for the reasoning path generator.}
    \label{fig:generation_pipeline}
\end{figure*}

\begin{figure*}[t]
    \centering
    \includegraphics[width=1\textwidth]{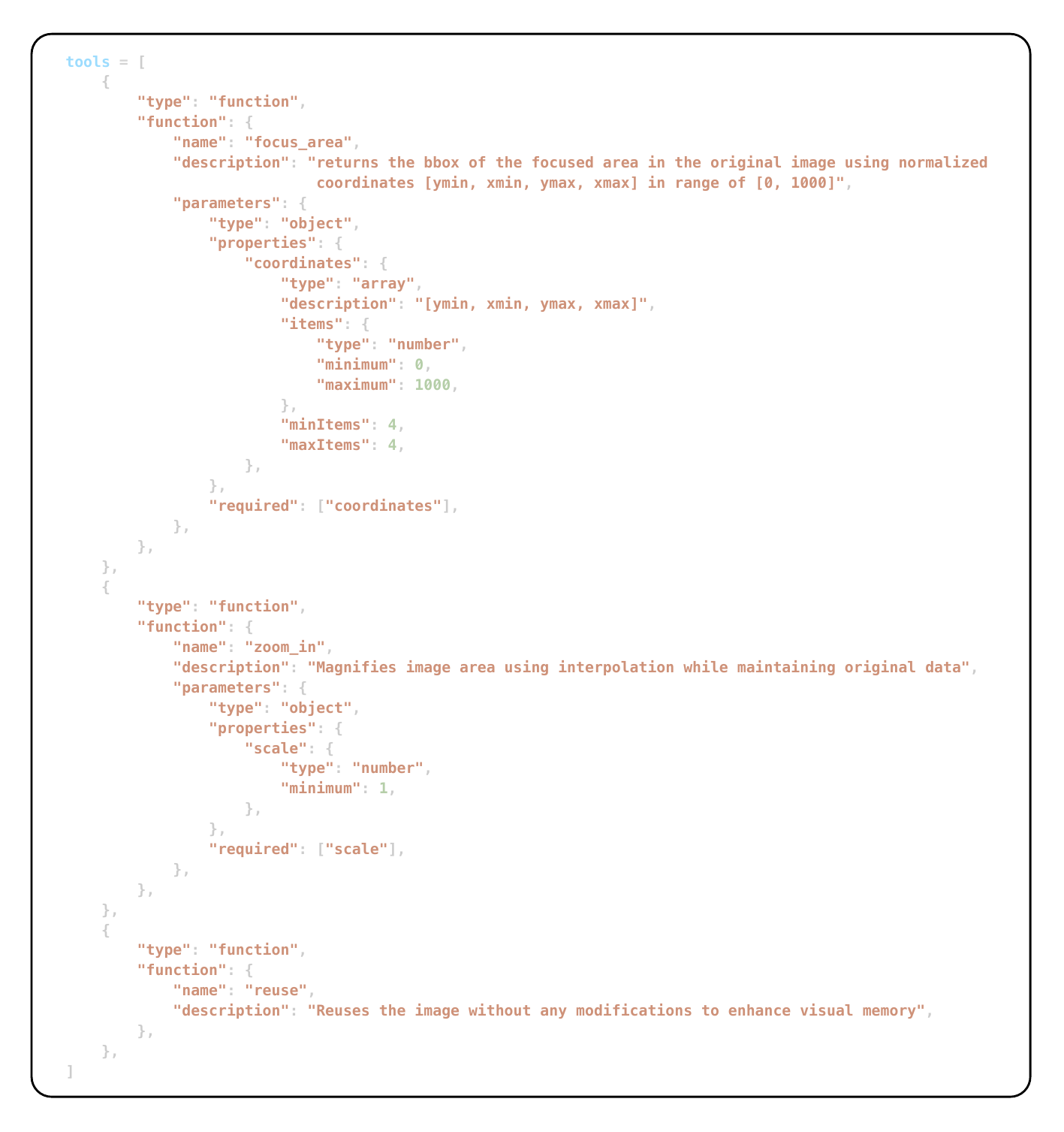}
    \caption{Tool definitions in OpenAI's dialogue format.}
    \label{fig:generation_pipeline}
\end{figure*}

\begin{figure*}[t]
    \centering
    \includegraphics[width=1\textwidth]{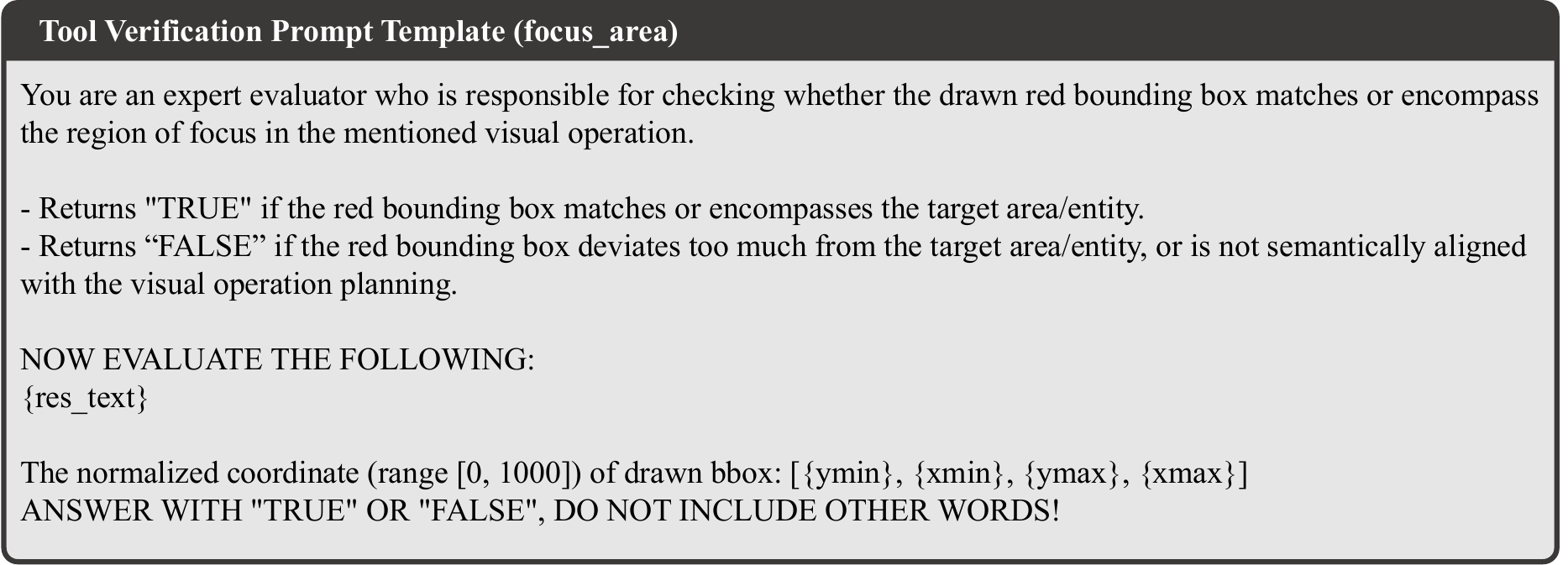}
    \caption{Tool verification prompt when executing $focus\_area$. The model determines whether the input image and the returned coordinates are semantically aligned with the visual operation planning.}
    \label{fig:tool_verification_focus_area}
\end{figure*}

\begin{figure*}[t]
    \centering
    \includegraphics[width=1\textwidth]{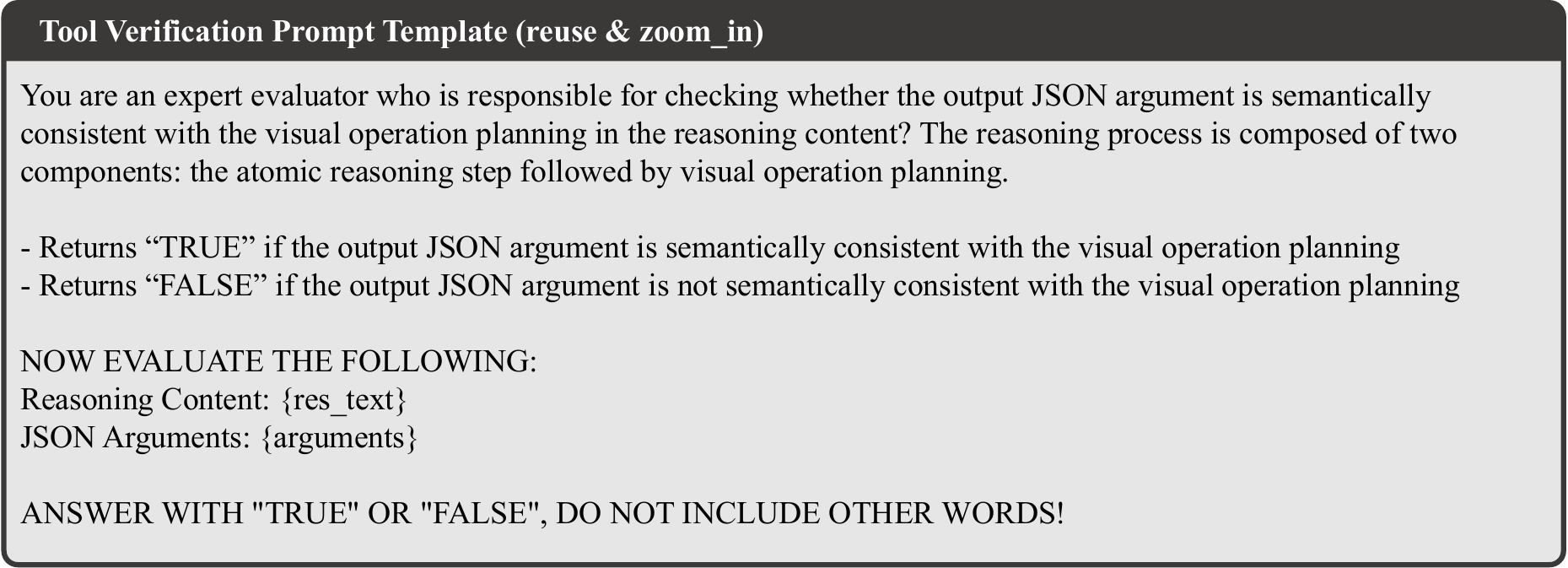}
    \caption{Tool verification prompt when executing $reuse$ and $zoom\_in$. The model determines whether the returned function is semantically aligned with the visual operation planning.}
    \label{fig:tool_verification_reuse_zoom}
\end{figure*}

\begin{figure*}[t]
    \centering
    \includegraphics[width=1\textwidth]{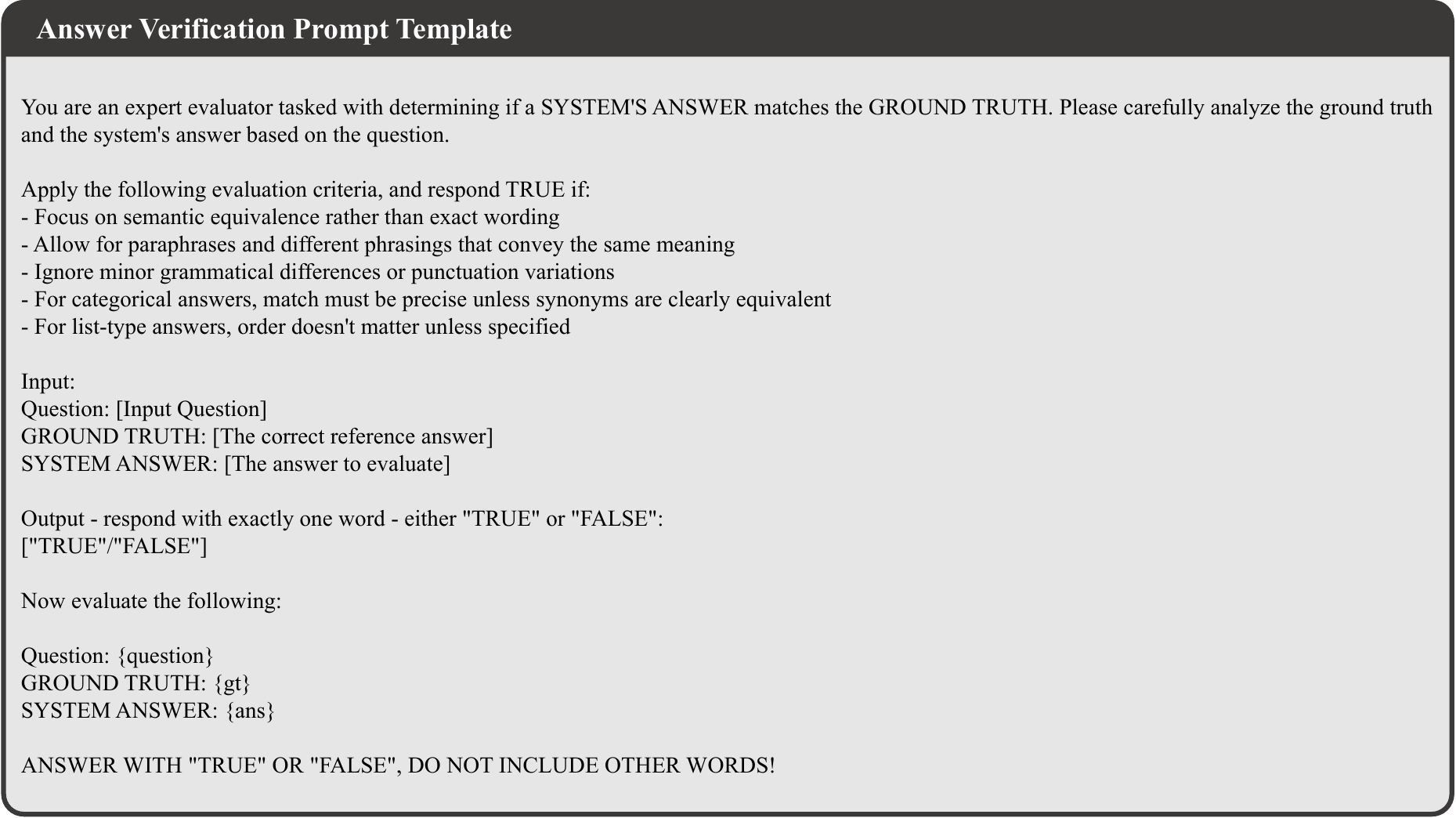}
    \caption{Answer verification prompt. The model judges whether the the system output is accurate based on the question and ground truth.}
    \label{fig:answer_verification}
\end{figure*}

\begin{figure*}[t]
    \centering
    \includegraphics[width=1\textwidth]{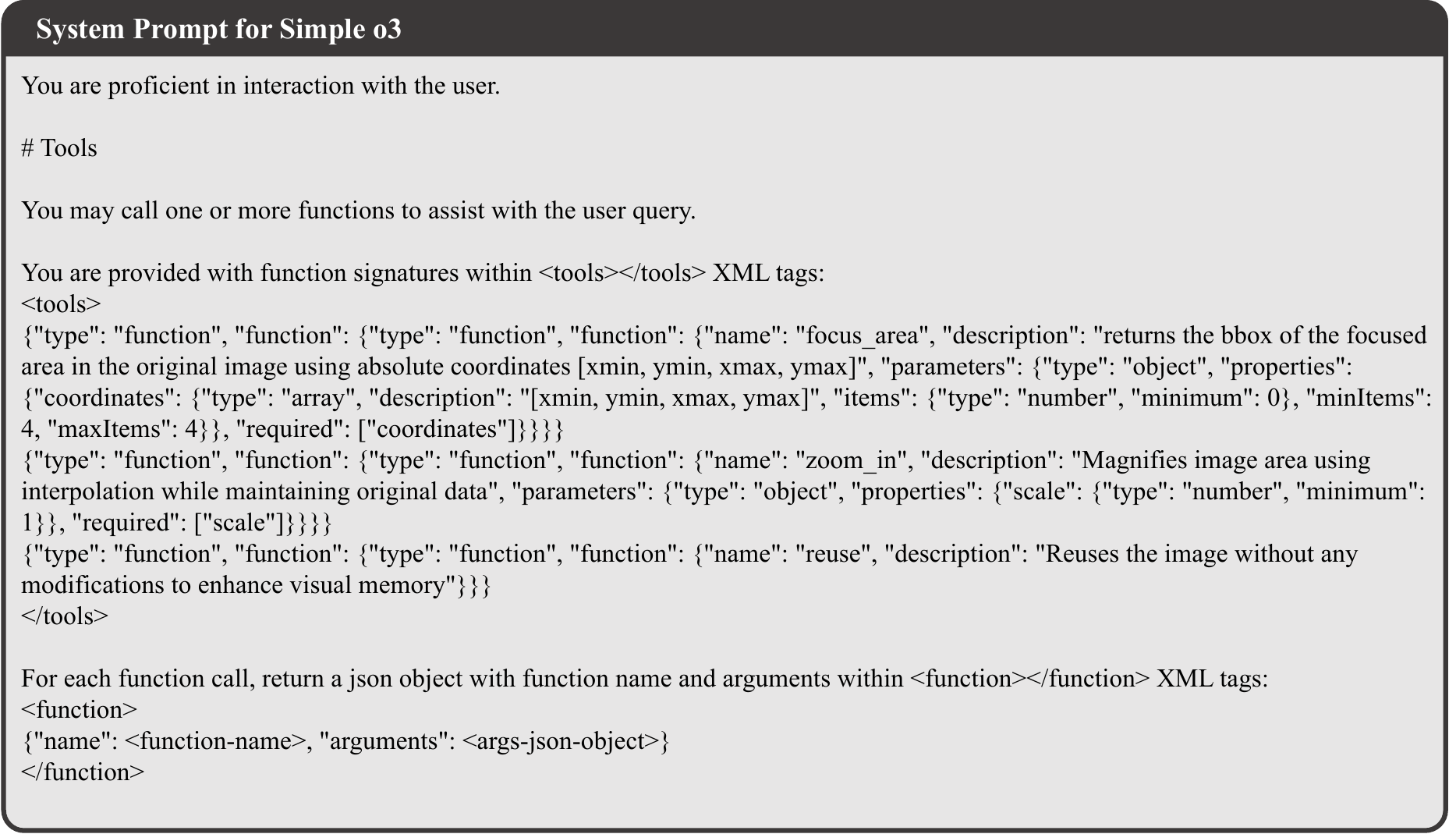}
    \caption{System prompt for Simple o3. Simple o3 calls the corresponding tool during the inference stage according to the definition, thereby operating the image to complete interleaved visual-language reasoning.}
    \label{fig:system_prompt_simple_o3}
\end{figure*}

\begin{figure*}[t]
    \centering
    \includegraphics[width=1\textwidth]{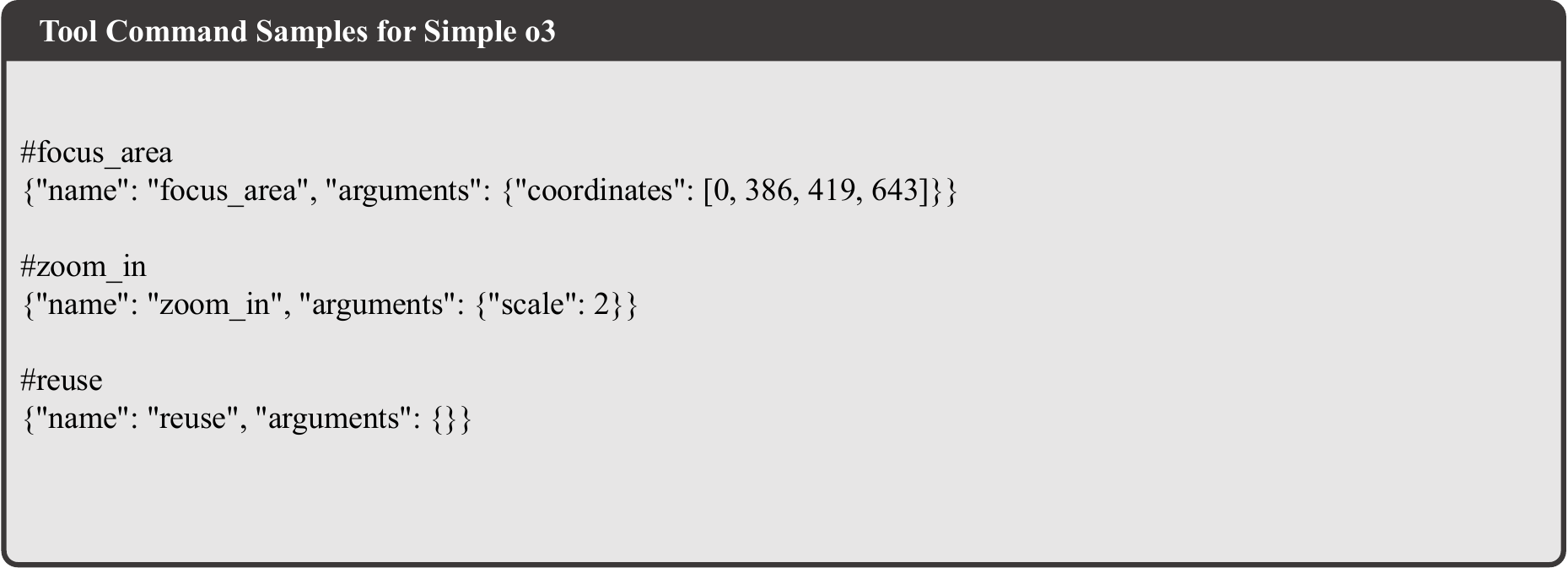}
    \caption{Examples of three tool commands in JSON format, including $focus\_area$, $zoom\_in$, and $reuse$.}
    \label{fig:tool_command}
\end{figure*}

\begin{figure*}[t]
    \centering
    \includegraphics[width=1\textwidth]{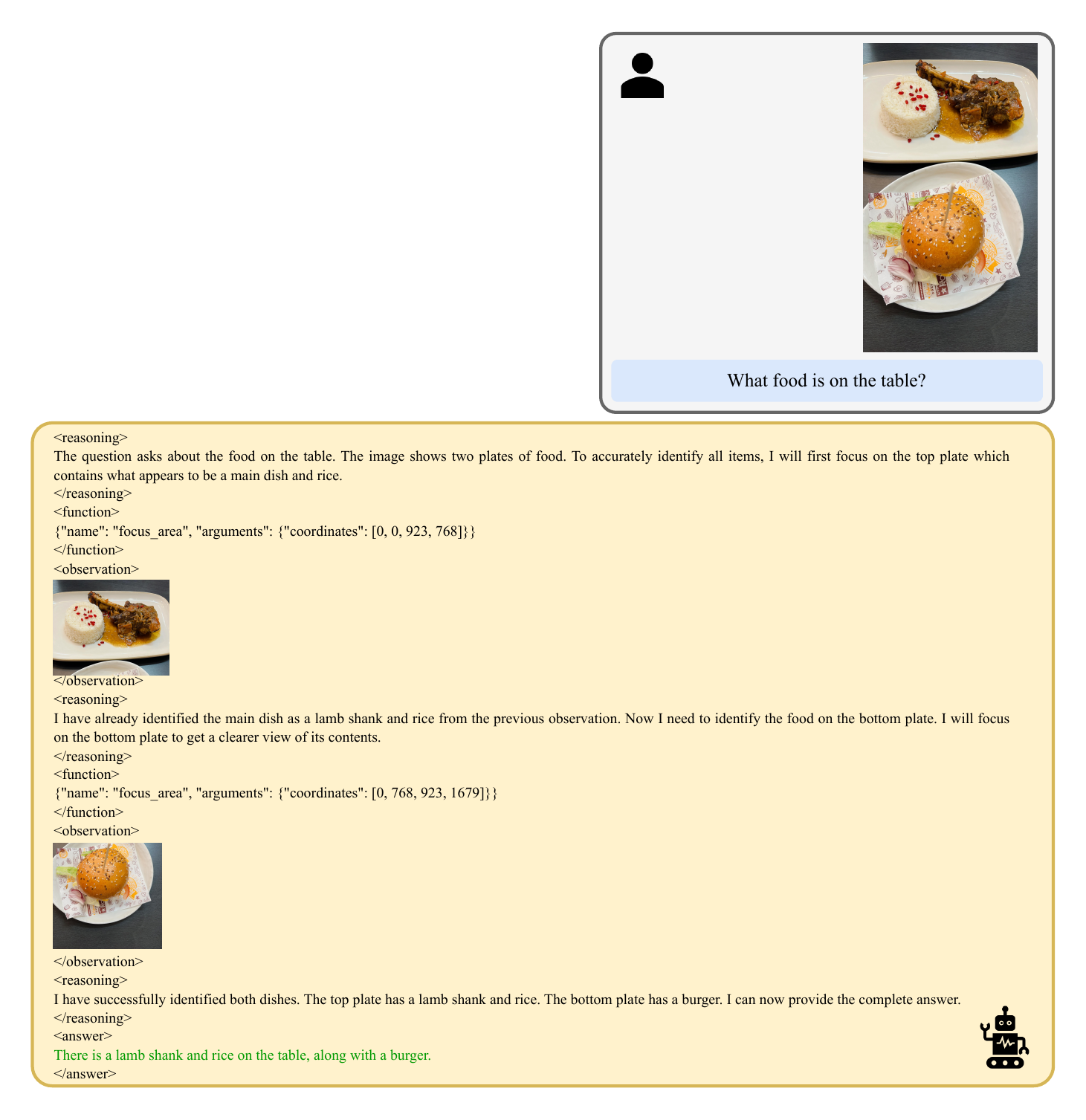}
    \caption{Multi-step interleaved vision-language reasoning of Simple o3 – Case 1. Simple o3 uses atomic reasoning steps to gradually analyze and accurately locate each entity.}
    \label{fig:case_1}
\end{figure*}

\begin{figure*}[t]
    \centering
    \includegraphics[width=1\textwidth]{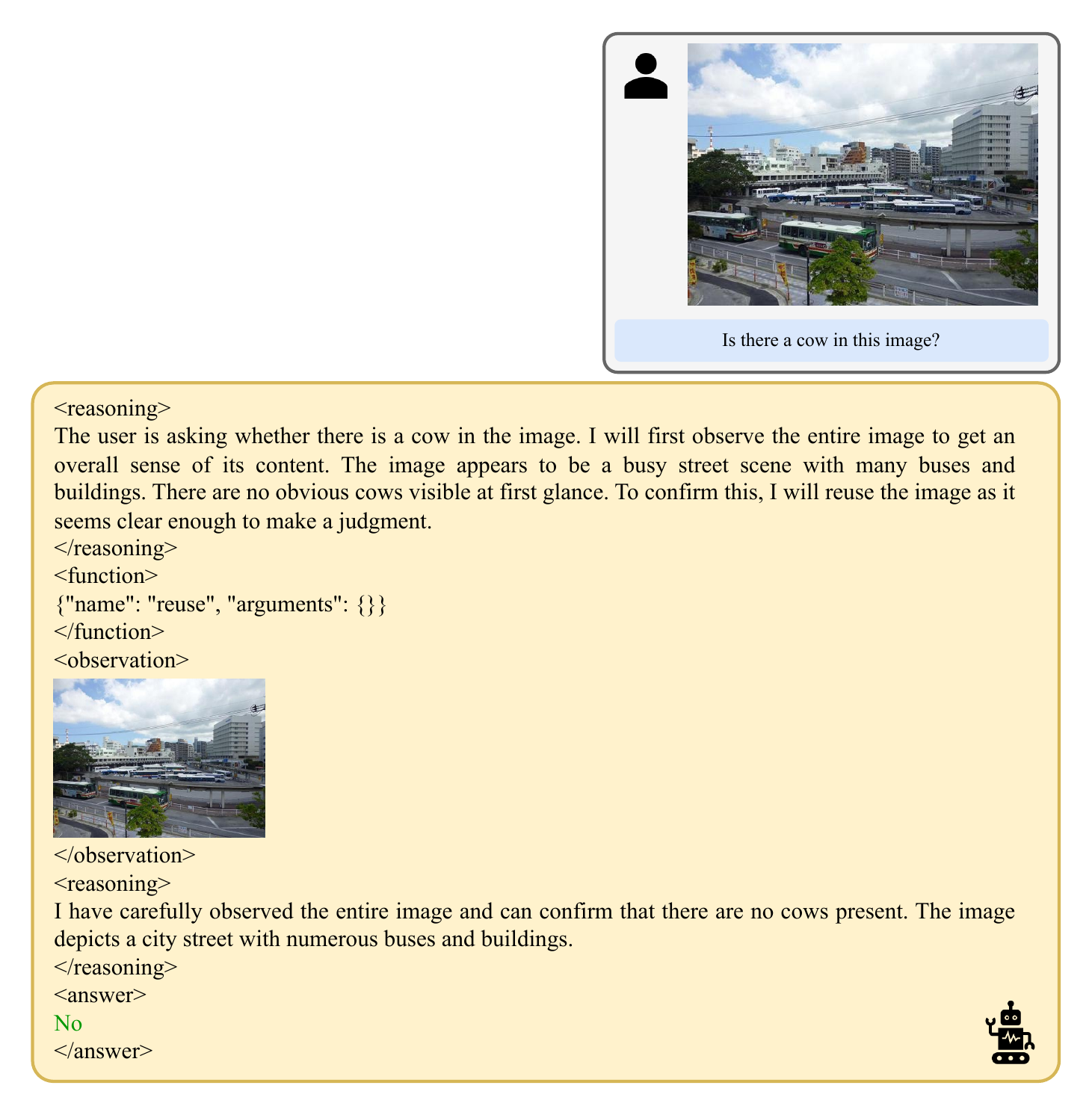}
    \caption{Multi-step interleaved vision-language reasoning of Simple o3 – Case 2. Simple o3 can reconfirm the deduction by reusing the original image.}
    \label{fig:case_2}
\end{figure*}

\begin{figure*}[t]
    \centering
    \includegraphics[width=1\textwidth]{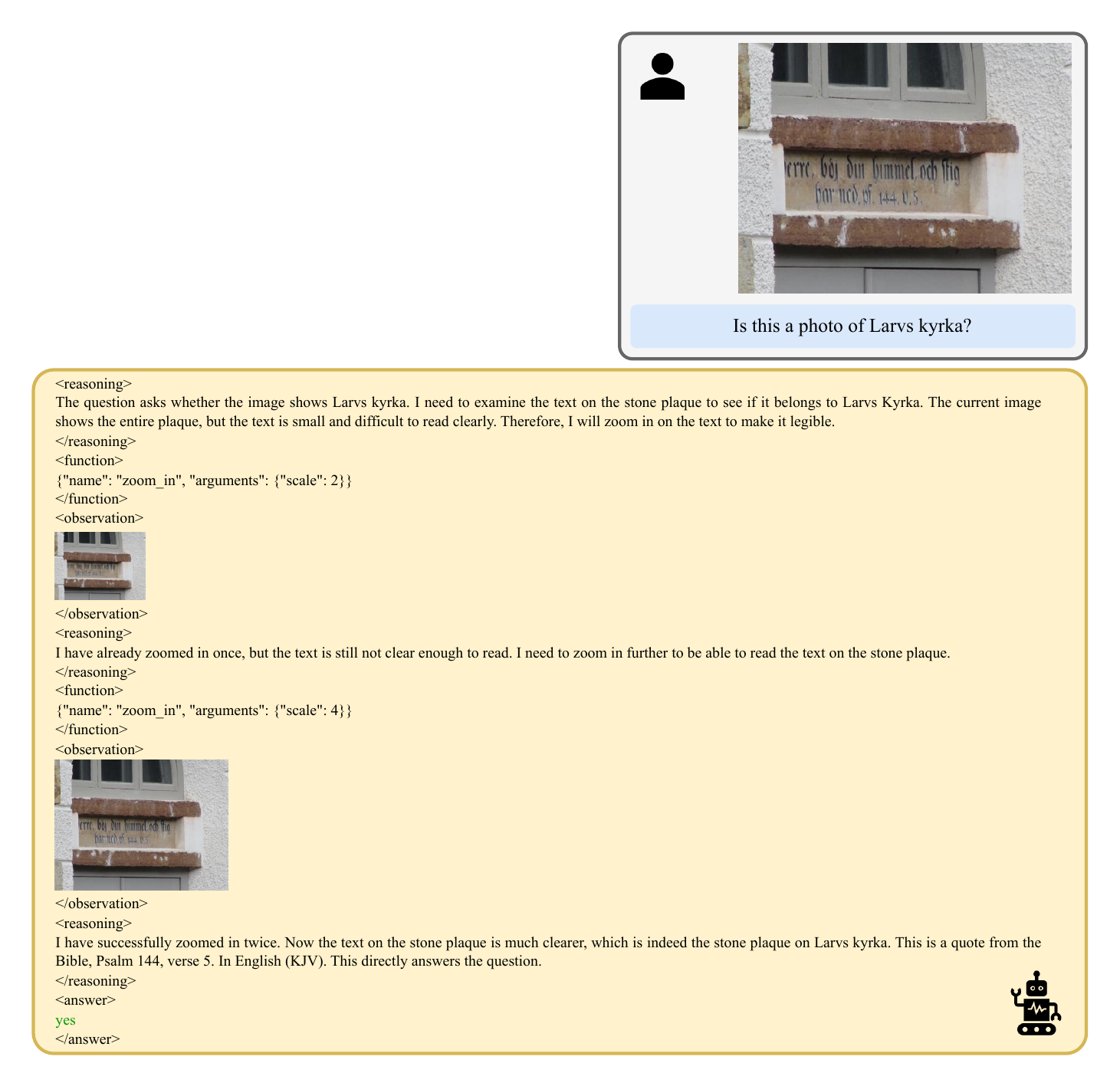}
    \caption{Multi-step interleaved vision-language reasoning of Simple o3 – Case 3. If Simple o3 detects that certain entities or fonts in an image are too small, it can enhance perception of the current image through iterative magnification.}
    \label{fig:case_3}
\end{figure*}

\begin{figure*}[t]
    \centering
    \includegraphics[width=1\textwidth]{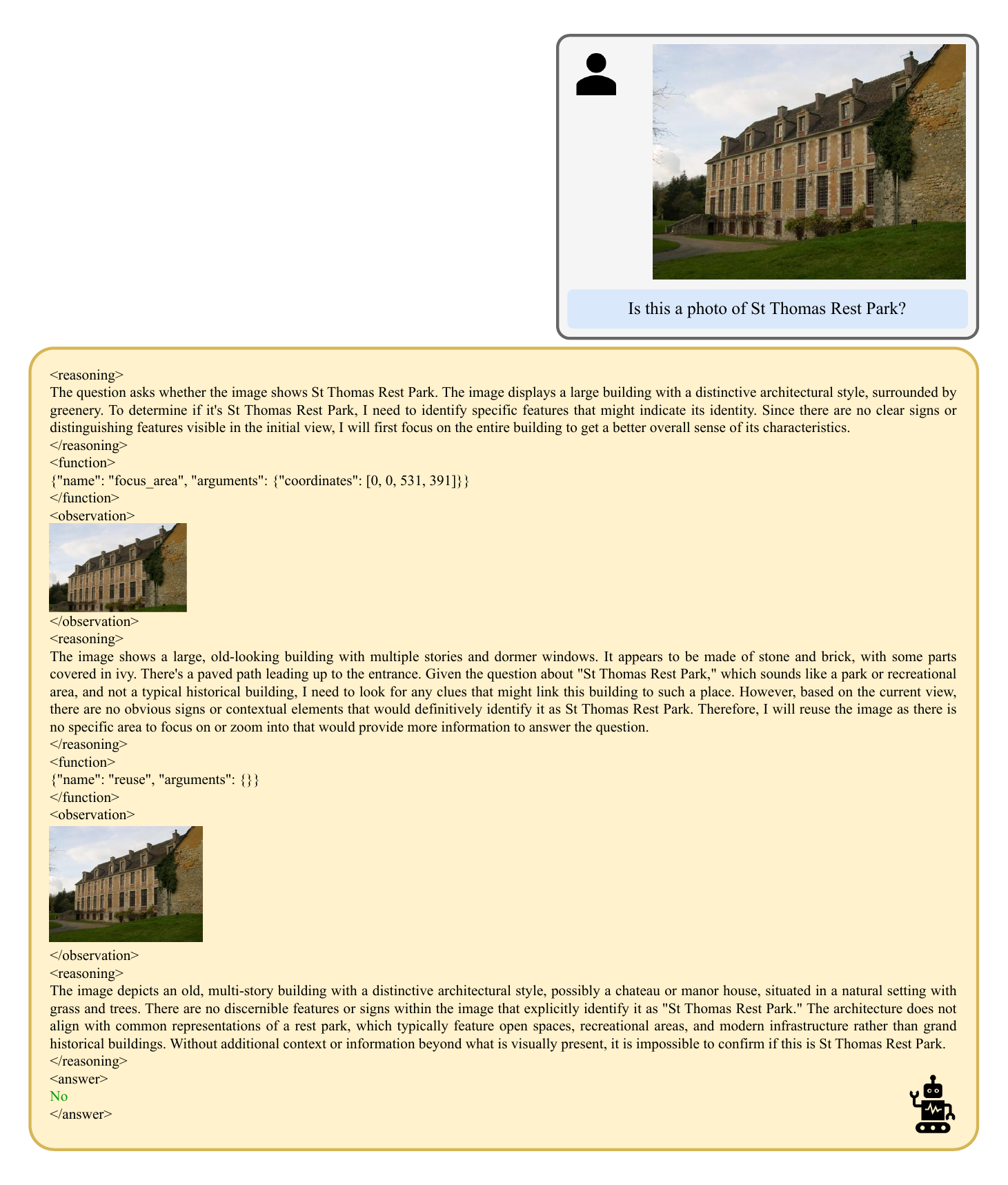}
    \caption{Multi-step interleaved vision-language reasoning of Simple o3 – Case 4. Simple o3 extends both text tokens and visual tokens simultaneously, further enhancing its reasoning capabilities.}
    \label{fig:case_4}
\end{figure*}

\end{document}